\documentclass[lettersize,journal]{IEEEtran}
\usepackage{array}
\usepackage[caption=false,font=footnotesize,labelfont=sf,textfont=sf]{subfig}
\usepackage{url}
\usepackage{booktabs}
\usepackage{colortbl}
\usepackage{graphicx}
\usepackage{amsmath}
\usepackage{amssymb}
\usepackage{cuted}
\usepackage{epsfig}
\usepackage{epstopdf}
\usepackage{makecell,multirow,diagbox}
\usepackage{color}
\usepackage{soul}
\usepackage{url}
\usepackage{amsmath}
\usepackage[utf8]{inputenc}
\usepackage{xcolor}
\usepackage{hyperref}
\usepackage{mathtools,xspace}
\usepackage{ragged2e}
\usepackage{times}
\hypersetup{
    colorlinks=false,
    breaklinks=false,
    linkcolor= red,
    bookmarksopen=false,
    filecolor=black,
    citecolor=blue,
    linkbordercolor=blue
}

\newcommand{\onedot}{\ifx\@let@token.\else.\null\fi\xspace}
\newcommand{\etal}{\emph{et al}\onedot}
\newcommand{\eg}{\emph{e.g}\onedot}
\newcommand{\ie}{\emph{i.e}\onedot}

\newcommand{\textred}{\textcolor[rgb]{1,0,0}}
\begin{document}

\title{Elite360M: Efficient 360 Multi-task Learning via Bi-projection Fusion and Cross-task Collaboration}

\author{
Hao Ai, \IEEEmembership{Student Member,~IEEE},
Lin Wang,~\IEEEmembership{Member,~IEEE}

\thanks{Corresponding author: Lin Wang (e-mail: linwang@ust.hk).}
\thanks{H. Ai is with the Artificial Intelligence Thrust, The Hong Kong University of Science and Technology (HKUST), Guangzhou, China. E-mail: hai033@connect.hkust-gz.edu.cn.}
\thanks{L. Wang is with the Artificial Intelligence Thrust, HKUST, Guangzhou, and Dept. of Computer Science and Engineering, HKUST, Hong Kong SAR, China. E-mail: linwang@ust.hk.}
}

\markboth{Journal of \LaTeX\ Class Files,~Vol.~14, No.~8, August~2021}%
{Shell \MakeLowercase{\textit{et al.}}: A Sample Article Using IEEEtran.cls for IEEE Journals}


\maketitle

\begin{abstract}
360$^\circ$ cameras capture the entire surrounding environment with a large field-of-view (FoV), exhibiting comprehensive visual information to directly infer the 3D structures, \eg, depth and surface normal, and semantic information simultaneously. Existing works predominantly specialize in a single task, leaving multi-task learning of 3D geometry and semantics largely unexplored. Achieving such an objective is, however, challenging due to: 1) inherent spherical distortion of planar equirectangular projection (ERP) and insufficient global perception induced by 360$^\circ$ image's ultra-wide FoV (360$^\circ$ $\times$ 180$^\circ$); 2) non-trivial progress in effectively merging geometry and semantics among different tasks to achieve mutual benefits.
In this paper, we propose a novel end-to-end multi-task learning framework, named \textit{Elite360M}, capable of inferring 3D structures via \eg, depth and surface normal estimation, and semantics via semantic segmentation simultaneously. Our key idea is to build a representation with strong global perception and less distortion while exploring the inter- and cross-task relationships between geometry and semantics. We incorporate the distortion-free and spatially continuous icosahedron projection (ICOSAP) points and combine them with ERP to enhance global perception. With a negligible cost (\textit{$\thicksim$1M parameters}), a Bi-projection Bi-attention Fusion (B2F) module is thus designed to capture the semantic- and distance-aware dependencies between each pixel of the region-aware ERP feature and the ICOSAP point feature set. Moreover, we propose a novel Cross-task Collaboration (CoCo) module to explicitly extract task-specific geometric and semantic information from the learned representation to achieve preliminary predictions. It then integrates the spatial contextual information among tasks to realize cross-task fusion.
Extensive experiments demonstrate the effectiveness and efficacy of Elite360M, outperforming the prior multi-task learning methods (designed for planar images) with significantly fewer parameters on two benchmark datasets. Moreover, our Elite360M exhibits on-par performance with the single-task learning methods. Code is available at \url{https://VLIS2022.github.io/Elite360M}.

\end{abstract}


\section{Introduction}
\label{sec:intro}
\begin{figure}[!t]
\centering
\includegraphics[width=1\linewidth]{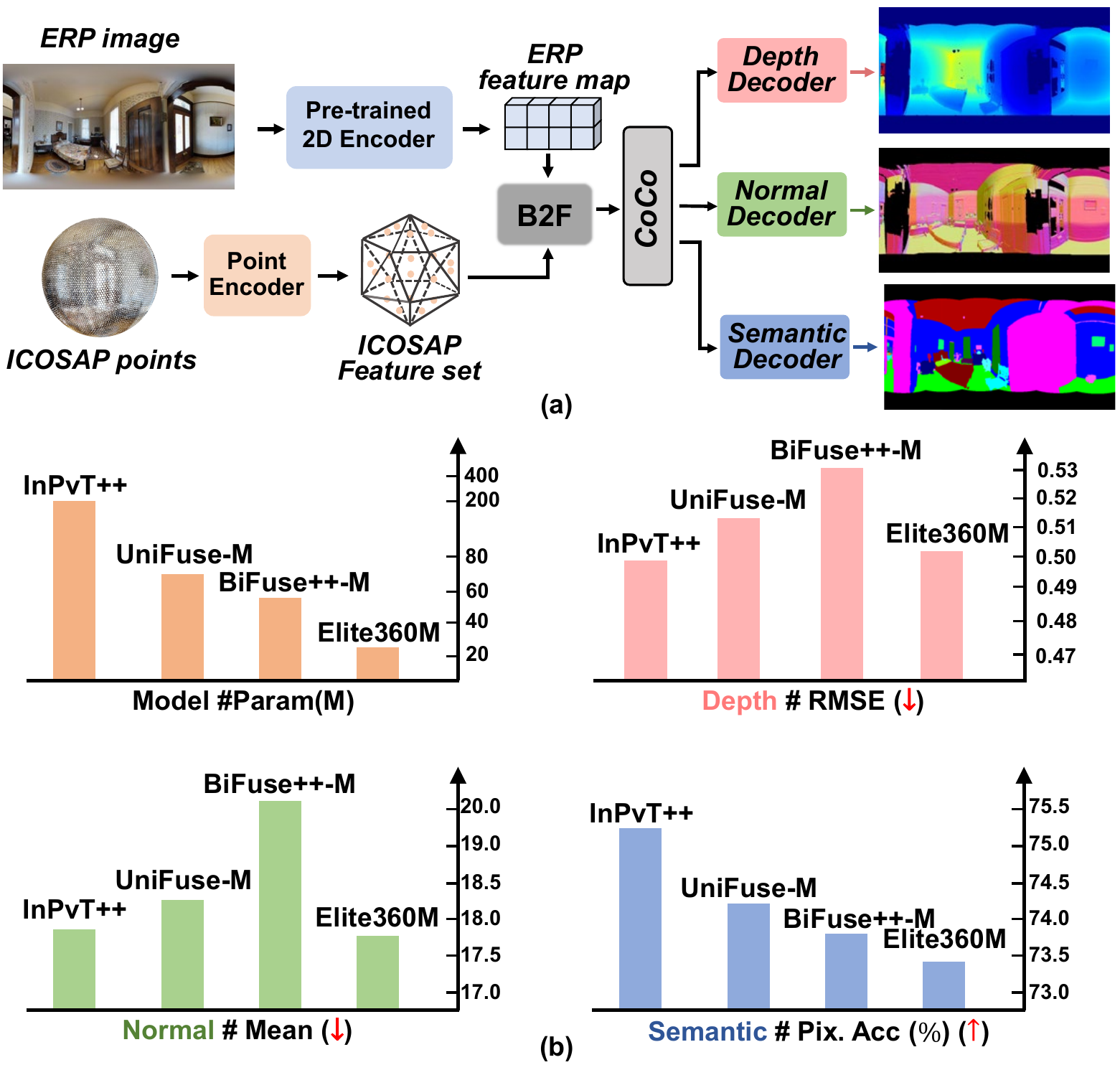}
\caption{(a) \textbf{Overview of Elite360M}. It employs the B2F module to learn the 360-specific representations from a local-with-global perspective and introduces CoCo module to model the cross-task information interaction. Consequently, with three simple decoder heads, Elite360M accomplishes three scene understanding tasks simultaneously. (b) \textbf{Performance of three tasks on Matterport3D test dataset~\cite{Chang2017Matterport3DLF}} (Root Mean Square Error (RMSE) for depth estimation, mean error of angles (Mean) for surface normal estimation, and pixel accuracy for semantic segmentation). Lower error or larger accuracy is better. Notably, UniFuse-M and BiFuse++-M are the multi-task learning frameworks built on UniFuse~\cite{Jiang2021UniFuseUF} and BiFuse++~\cite{Wang2022BiFuseSA}, respectively. We follow the original decoder structures to build three decoder branches for three tasks. InPvT++ is the SOTA multi-task learning method for conventional planar images. In particular, UniFuse-M, BiFuse++-M and our Elite360M are with the ResNet-34 as the ERP encoder backbone, while InPvT++ employs much larger ViT-Large~\cite{dosovitskiy2021an} as the backbone.}
\label{fig:coverfig}
\vspace{-10pt}
\end{figure}

\IEEEPARstart{T}{h}e 360$^\circ$ images\footnote{360$^\circ$ images are also called panoramas or omnidirectional images.} are captured with a wide field-of-view (FoV) of $360^\circ \times 180^\circ$ and record complete surrounding scene details~\cite{Barron2022MipNeRF3U, daSilveira20223DSG, Ai2022DeepLF,Wang2022BiFuseSA,Zhang2022BehindED}. 360$^\circ$ cameras enjoy broad applications for holistic scene understanding in diverse fields, such as autonomous driving, virtual reality, and robotic navigation.
Compared to the planar images, a single 360$^\circ$ image provides comprehensive visual information to directly infer the 3D structures, \eg, depth and surface normal, and semantic information simultaneously. 
This highlights a promising research direction of learning multiple scene understanding tasks, if possible, with one versatile model. This not only significantly enhances efficiency but also facilitates mutual assistance through the relationships between the 3D geometry and semantics~\cite{Crawshaw2020MultiTaskLW,Vandenhende2020MultiTaskLF,Bhattacharjee2022MuITAE,invpt2022,Ye2023InvPTIP}. 

To date, most dominant research works for 360$^\circ$ images focus on accomplishing a single task in one training iteration, either depth estimation~\cite{Jiang2021UniFuseUF,Wang2020BiFuseM3,Wang2022BiFuseSA,Ai2023HRDFuseM3} or normal estimation~\cite{Karakottas2019360SR}, or semantic segmentation~\cite{Li2023SGAT4PASSSG,Yang2021CapturingOC}. It still remains an under-explored for multi-task learning based on a single 360$^\circ$ image. There are two viable options to achieve 360 multi-task learning of 3D geometry and semantics, which simultaneously predicts the above scene understanding tasks with one versatile model. One is to adapt existing multi-task learning networks designed for planar images, \eg,~\cite{invpt2022,Ye2023InvPTIP}, to train on planar projections of 360$^\circ$ images, and another is to design the 360-specific multi-task learning network.

\begin{figure}[!t]
\centering
 \includegraphics[width=0.95\linewidth]{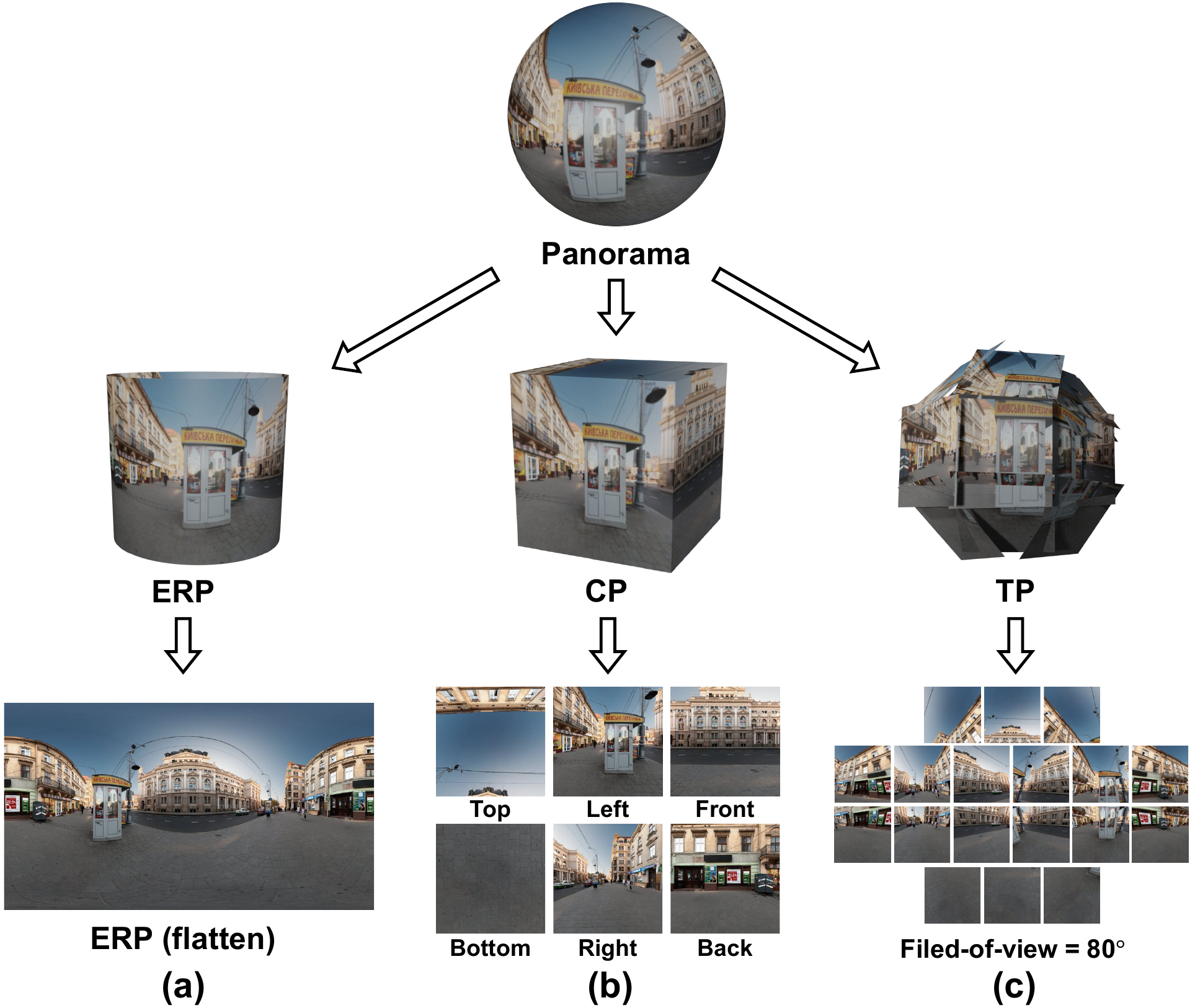}
\caption{Different planar projections of a spherical imaging panorama: a) equirectangular projection (ERP); b) cubemap projection (CP); c) tangent projection (TP) (captured following~\cite{Li2022OmniFusion3M}) with the FoV of $(80^\circ,80^\circ)$.}
\label{fig:projection}
\vspace{-15pt}
\end{figure}

To adapt conventional planar methods to 360$^\circ$ images, the prevalent practice is to train these networks on the most commonly used planar projection data, \ie, equirectangular projection (ERP). However, as shown in Fig.~\ref{fig:projection}(a), projecting raw spherical data onto 2D planes results in severe distortion in ERP images, particularly at the poles. Moreover, understanding the ultra-wide FoV content in a 360$^\circ$ image necessitates robust global perception capabilities. This makes it difficult to be processed by the limited local receptive fields of convolutional filters and fixed window sizes of local attention mechanisms\cite{Ai2024Elite360DTE,Yun2023EGformerEG}.
These two issues result in sub-optimal results when adapting the methods designed for planar images. For instance, InvPT++~\cite{Ye2023InvPTIP} incurs high computational costs and yields only satisfactory performance, as illustrated in Fig.~\ref{fig:coverfig}(b)). When designing a multi-task learning framework for 360$^\circ$ images, it is essential to not only address the two aforementioned issues but also to effectively merge geometry and semantics across different tasks to promote mutual enhancement---a crucial aspect that is often overlooked. The latest work, MultiPanoWise~\cite{shah2024multipanowise}, employs PanoFormer~\cite{Shen2022PanoFormerPT} as the backbone and follows task-specific output heads to predict various tasks separately. However, it ignores the dependencies among different tasks and avoids cross-task information sharing, leading to less accurate results (See Tab.~\ref{tab:com_multi_matt}).

In this work, we propose a novel multi-task learning framework, named \textbf{Elite360M}, aiming to simultaneously learn the 3D geometry via depth and surface normal estimation and semantics via semantic segmentation. The key idea is to \textit{build a representation with strong global perception and less distortion while exploring the inter- and cross-task relationships between geometry and semantics}. Our Elite360M includes two key components: the bi-projection fusion facilitated by an efficient Bi-projection Bi-attention Fusion (\textbf{B2F}) module and the attention-based Cross-task Collaboration (\textbf{CoCo}) module aiming to capture the cross-task interactions. 

\begin{figure}[!t]
\centering
\includegraphics[width=0.98\linewidth]{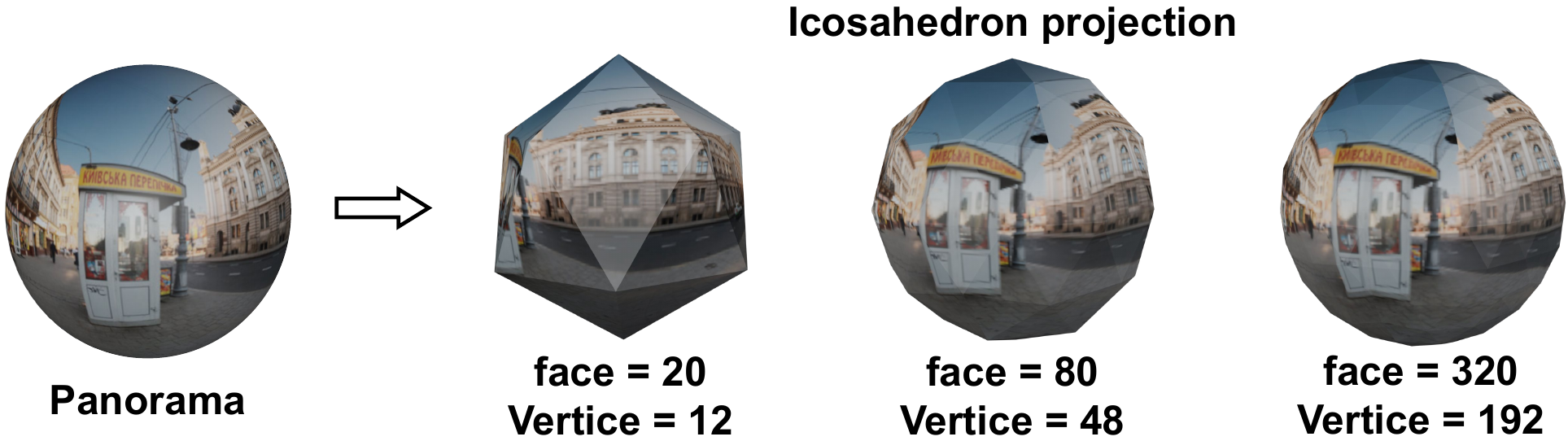}
\caption{Icosahedron projection (ICOSAP) with different subdivision levels.}
\label{fig:icosap}
\vspace{-15pt}
\end{figure}

Concretely, our B2F module incorporates the spatially continuous and distortion-free non-Euclidean projection $\textendash$ icosahedron projection (ICOSAP) points and combines them with ERP to enhance global perception, as depicted in Fig.~\ref{fig:icosap}. 
Bi-projection fusion has been widely applied to depth estimation~\cite{Wang2020BiFuseM3,Wang2022BiFuseSA,Jiang2021UniFuseUF} 
and capitalizes on the benefits of spatially complete ERP and other less-distorted planar projections (cubemap projection (CP)~\cite{Cheng2018CubePF} (Fig.~\ref{fig:projection}b) and tangent projection (TP)~\cite{Eder2019TangentIF} (Fig.~\ref{fig:projection}c)). For example, BiFuse~\cite{Wang2020BiFuseM3, Wang2022BiFuseSA} and UniFuse~\cite{Jiang2021UniFuseUF} fuse ERP feature maps with cubemap-to-ERP feature maps. Meanwhile, HRDFuse~\cite{Ai2023HRDFuseM3} integrates patch-wise TP predictions into an ERP format depth map and fuses it with the depth map predicted from the ERP input image. Nevertheless, they remarkably increase computational costs (See Tab.~\ref{tab:com_single_matt}) and are hardly possible to provide global perception capabilities, as each ERP pixel aligns with a TP/CP patch in a one-to-one manner. Our Elite360M employs ICOSAP and represents the spherical data as discrete point sets rather than icosahedron meshes~\cite{Zhang2019OrientationAwareSS,Cohen2019GaugeEC,Jiang2019SphericalCO} or unfolded representations~\cite{Lee2018SpherePHDAC,Yoon2021SphereSRI} (See Fig.~\ref{fig:icosap}). As such, it enables the avoidance of semantic information redundancy due to pre-existing dense ERP pixels and the preservation of crucial continuous positional information from ICOSAP. 
\textit{Optimally, our B2F module captures the semantic- and distance-aware dependencies between each ERP pixel feature and the entire ICOSAP feature set}. This representation learned from the combination of ERP and ICOSAP establishes the foundational prerequisites for learning 3D geometry and semantics simultaneously. 

On top of the B2F module, we also introduce an attention-based CoCo module, that subtly leverages the inter- and cross-task relationships to merge the geometric and semantic information from different tasks, inspired by previous methods for planar images~\cite{Xu2022MTFormerML,Ye2023InvPTIP}.
CoCo module consists of two parts, preliminary predictions and cross-task fusion. Firstly, before cross-task interactions, we follow~\cite{invpt2022,Ye2023InvPTIP} to introduce the preliminary predictions for \textit{explicitly disentangling the task-specific information from the shared representation}. As shown in Fig.~\ref{fig:coco}(a), without complex network designs, preliminary predictions simply use ground truth labels to supervise the extraction of task-specific features, ensuring that each branch's features contain only information relevant to the corresponding task. Subsequently, the cross-task attention mechanism \textit{establishes strong dependencies across tasks by sharing the attention messages between different tasks}. Specifically, considering that depth estimation is closely related to 3D structure and semantics, we share the spatial contextual information in the depth features with the normal and semantic features for beneficial cross-task interactions. This approach reduces redundant computation of the intricate information exchange among the three tasks and greatly improves the performance (See Tab.~\ref{tab:ab_coco_part}).

\begin{figure*}[!t]
\centering
\includegraphics[width=\textwidth]{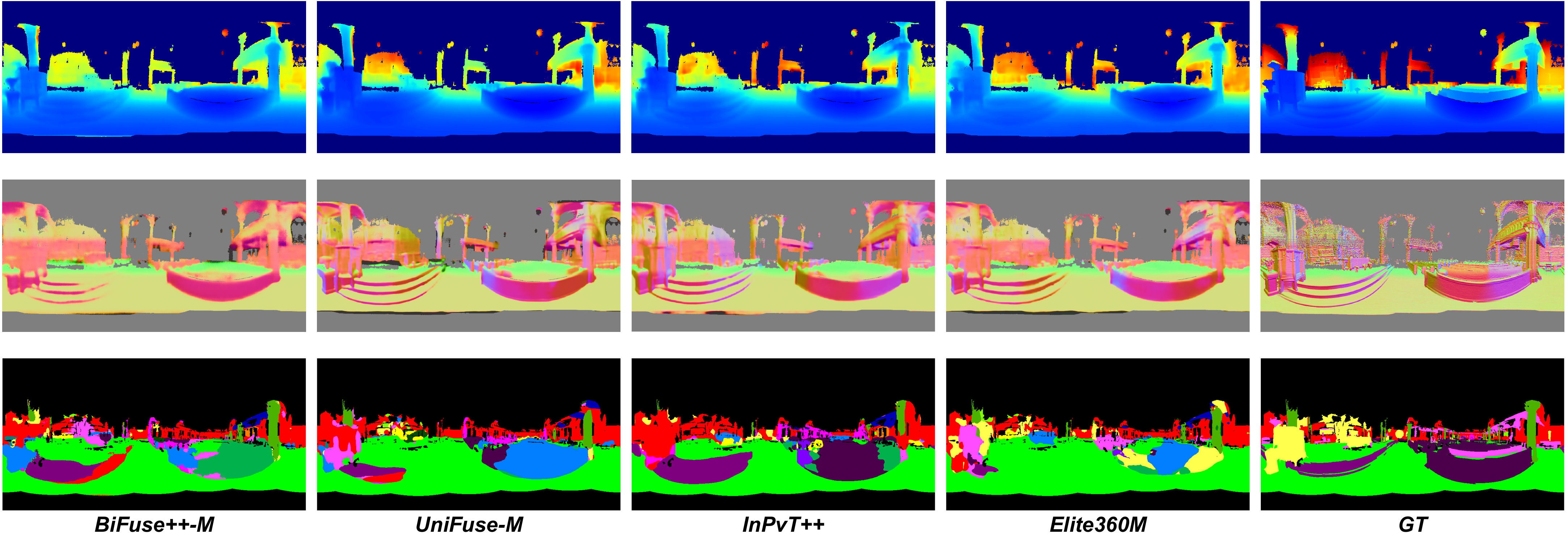}
\vspace{-12pt}
\caption{Qualitative comparison of one test sample of Matterport dataset among Elite360M and other multi-task learning baselines.}
\vspace{-12pt}
\label{fig:com_multi}
\end{figure*}

We evaluate the performance of Elite360M on two large-scale and challenging multi-task scene understanding benchmarks: the real-world Matterport3D dataset~\cite{Chang2017Matterport3DLF} and synthetic Structured3D dataset~\cite{Zheng2019Structured3DAL}. Our comprehensive experiments across various tasks show that Elite360M not only outperforms several popular multi-task learning baselines with much lower computational costs but also achieves performance comparable to single-task learning methods with even fewer parameters. Additionally, we provide extensive ablation analyses to demonstrate the crucial role of the B2F and CoCo modules.

This manuscript presents a substantial improvement over our CVPR 2024 work~\cite{Ai2024Elite360DTE}, achieved by extending the method and experiments in five aspects: \textbf{\uppercase\expandafter{\romannumeral1}}) We propose a novel end-to-end multi-task learning framework, \textit{\textbf{Elite360M}}, which leverages a learned distortion-aware and globally perceived representation to simultaneously infer 3D structures (\ie, depth and surface normal) and semantic information; \textbf{\uppercase\expandafter{\romannumeral2}}) We design an efficient and effective CoCo module to extract task-specific information for each task and exploit the inter- and cross-task relationships to aggregate spatial contextual information across different tasks and improve multi-task learning performance; \textbf{\uppercase\expandafter{\romannumeral3}}) We extensively evaluate the performance of Elite360M on two challenging benchmarks, considering various multi-task learning baselines, single-task learning methods, and ERP encoder backbones; \textbf{\uppercase\expandafter{\romannumeral4}}) We provide substantial qualitative and quantitative results, along with ablation analyses, to further demonstrate the effectiveness and efficiency of our approach. The rest of the paper is organized as follows: Sec.~\ref{sec:relatedworks} reviews the related works, and Sec.~\ref{sec:method} provides a detailed introduction to the proposed Elite360M. Sec.~\ref{sec:experiment} presents the experimental results to validate the effectiveness of our proposal and discusses its limitations. Sec.~\ref{sec:conclusion} concludes the paper and outlines potential future directions

\section{Related Work}
\label{sec:relatedworks}

\subsection{360 Scene Understanding}

\subsubsection{360 Depth Estimation}

As the most popular and crucial task in 360 vision, 360 depth estimation has been the focus of extensive research. According to the input, previous methods can be categorized into two types:

\textbf{Single-projection input:} The mainstream approach uses ERP images as input. As a sphere-to-plane projection, ERP exhibits horizontal distortion and increased curvature compared to planar images, particularly in the polar regions. To alleviate this distortion, numerous sphere-aware convolution filters~\cite{Tateno2018DistortionAwareCF,Cheng2020OmnidirectionalDE,zioulis2018omnidepth,Zhuang2021ACDNetAC} and attention windows~\cite{Shen2022PanoFormerPT,Yun2023EGformerEG} have been proposed. The pioneering work, OmniDepth~\cite{zioulis2018omnidepth} uses row-wise rectangular filters to address distortion across different latitudes. Alternatively,\cite{Tateno2018DistortionAwareCF} and\cite{coors2018spherenet} adaptively adjust the sampling grids of convolution filters based on their locations in ERP images, employing distortion-aware convolution filters. Building on this distortion-aware sampling strategy, PanoFormer~\cite{Shen2022PanoFormerPT} introduces a pixel-wise deformable attention mechanism, positioning its attention windows on the sphere's tangent grids. Most recently, EGFormer~\cite{Yun2023EGformerEG} proposes vertical and horizontal attention windows with spherical position embedding. In addition to refining convolutional kernels or attention windows, some methods attempt to mitigate distortion effects at the input level. For example, SliceNet~\cite{Pintore2021SliceNetDD} and PanelNet~\cite{yu2023panelnet} partition an ERP image into vertical slices, apply standard convolutional layers to predict the depth map of each slice, and then reassemble them into the ERP format. Another approach involves using less-distorted projection data as inputs, such as CP patches and TP patches. 360MonoDepth~\cite{ReyArea2021360MonoDepthH3} and OmniFusion~\cite{Li2022OmniFusion3M} employ established off-the-shelf depth estimation networks, designed for planar images, to predict depth outcomes for each TP patch. The results are then re-projected onto the ERP format through a neural network.

\subsubsection{360 Semantic Segmentation}
Advances in deep learning have facilitated the development of 360 semantic segmentation~\cite{Tateno2018DistortionAwareCF, Zhuang2021ACDNetAC, Lee2018SpherePHDAC, Jiang2019SphericalCO, Sun2020HoHoNet3I, Zhang2022BehindED, Li2023SGAT4PASSSG}. The existing solutions focus on addressing the spherical distortion problem, or learning rotation-equivariant representations. For distortion-aware strategies, similar to the aforementioned CNN-based depth estimation methods, Yang~\etal~\cite{Yang2021CapturingOC} proposed Horizontal Segment Attention module to capture contextual information from the horizontal segments of ERP feature maps, which can reduce the effects of distortion. Meanwhile Hu~\etal~\cite{Hu2022DistortionCM} proposed a distortion-aware convolution network to correct the distortion according to the spherical imaging property. Besides, some transformer-based works~\cite{Zhang2022BehindED,Zhang2022BendingRD,Zheng2023LookAT,Li2023SGAT4PASSSG} replace the vanilla patch embeddings and Multi-Layer Perception (MLP) layers in the traditional transformer blocks with spherical geometry-based deformable patch embeddings and deformable MLP layers to accommodate the pixel densities of ERP images. For rotation-aware strategies, most methods~\cite{Lee2018SpherePHDAC,Zhang2019OrientationAwareSS,Cohen2019GaugeEC,esteves2020spin,shakerinava2021equivariant,Ocampo2022ScalableAE} represent 360$^\circ$ images as pixelizations of platonic solids, \eg, icosahedron, and propose non-Euclidean convolution operations to obtain the rotation-equivariant representations. However, these methods suffer from low resolution and high computational cost. In contrast, our bi-projection fusion represents ICOSAP data as the point sets and combines ICOSAP points with ERP images. On the top of the dense RGB information and pixel locations from ERP images, Elite360M achieves high-resolution semantic segmentation results with a low computational cost.

\subsubsection{360 Surface Normal Estimation}
Surface normal prediction is intrinsically linked to the 3D geometry of scenes. With the advancements in learning-based methods within computer vision, a variety of network architectures, ranging from CNNs~\cite{Huang2019FrameNetLL,Eftekhar2021OmnidataAS,Bae2024RethinkingIB} to vision transformers~\cite{Yang2021TransformerBasedAN,Kar20223DCC}, have been developed for planar images. However, the large field-of-view (FoV) content and intricate structural details of 360$^\circ$ environments pose significant challenges for surface normal estimation~\cite{Karakottas2019360SR,feng2020deep}. For instance, HyperSphere~\cite{Karakottas2019360SR} employs a standard CNN trained with a specialized loss derived from the quaternion product of estimated and ground-truth normal vectors. Similarly, Feng~\etal~\cite{feng2020deep} introduced a network that jointly trains for depth and surface normal estimation, utilizing a double quaternion approximation to formulate the supervision loss in a 4D hyper-spherical space. In this work, we aim to achieve end-to-end 360 multi-task learning that simultaneously addresses three scene understanding tasks—depth estimation, semantic segmentation, and surface normal estimation—leveraging the representation learned from ICOSAP-based bi-projection fusion and attention-driven cross-task interactions.

\vspace{-5pt}
\subsection{Multi-task learning}

The classic pipeline of a multi-task learning method, as demonstrated in~\cite{Eigen2014PredictingDS, Misra2016CrossStitchNF, Liu2018EndToEndML, Bhattacharjee2022MuITAE, Xu2022MTFormerML, Ye2023InvPTIP, Agiza2024MTLoRAAL}, involves first extracting a task-agnostic representation from the input using a shared encoder, followed by employing a set of task-specific decoders to generate predictions for each distinct task. Multi-task learning offers advantages over training separate models for each task, such as faster training and inference times and enhanced performance through the sharing of domain information between complementary tasks. According to the strategies of information sharing, multi-task learning can be categorized into two main approaches: hard parameter sharing and soft parameter sharing. Hard parameter sharing, employed in works such as ~\cite{Eigen2014PredictingDS, Bhattacharjee2022MuITAE, Ye2023InvPTIP}, involves sharing the lower layers across all tasks while keeping the upper and output layers task-specific. Conversely, soft parameter sharing, as seen in \cite{Misra2016CrossStitchNF, Liu2018EndToEndML, Agiza2024MTLoRAAL}, establishes separate backbones for each task and integrates inter-task information among these backbones, enhancing cross-task synergy\cite{Crawshaw2020MultiTaskLW, Vandenhende2020MultiTaskLF}.

Although multi-task learning for planar images has achieved significant success, its application to 360$^\circ$ images remains largely unexplored. Zeng~\etal~\cite{Zeng2020Joint3L} proposed a joint learning method that leverages the consistency of structural information between room layout and depth estimation for mutual refinement. Pano-SfMLearner~\cite{Liu2021PanoSfMLearnerSM} integrates depth and pose estimation to generate novel views, enhancing segmentation results through cross-view loss functions. The recent MultiPanoWise~\cite{shah2024multipanowise} initially employs a hard parameter sharing strategy for multi-task predictions, introducing a refinement stage to utilize features from the encoding process to optimize initial predictions. However, these approaches do not fully explore and utilize consistent geometric and semantic information across different tasks to mutually enhance performance. Inspired by methodologies designed for planar images~\cite{Ye2023InvPTIP, Bhattacharjee2022MuITAE}, we introduce the CoCo module, which effectively disentangles task-specific features from task-agnostic representations and facilitates beneficial cross-task message passing.
\begin{figure*}[t!]
\centering
\includegraphics[width=0.97\textwidth]{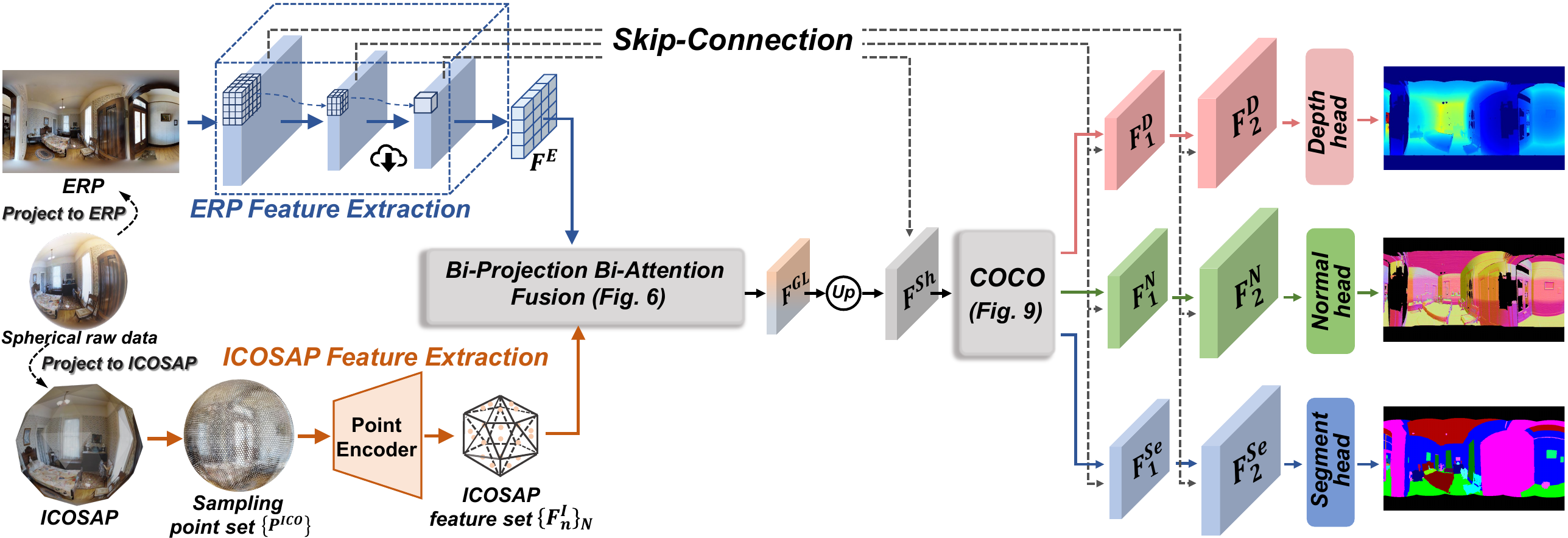}
\caption{An overview of our \textbf{\textit{Elite360M}} framework, comprising image-based ERP feature extraction (Sec.~\ref{sec:erp_feature}), point-based ICOSAP feature extraction (Sec.~\ref{sec:ico_feature}), Bi-projection Bi-attention fusion (B2F) (Sec.~\ref{sec:b2f}), Cross-task Collaboration (CoCo) (Sec.~\ref{sec:coco}), and three task-specific decoders (Sec.~\ref{sec:loss}). Notably, We employ the skip connections~\cite{ronneberger2015u} at the decoding stage.}
\label{fig:overview}
\vspace{-10pt}
\end{figure*}

\vspace{-15pt}
\subsection{Cross-Attention Mechanism}
Cross-attention is widely utilized for efficient multi-modal feature fusion. For example, Chen~\etal~\cite{Chen2022AutoAlignPF} developed cross-attention blocks to align and fuse 2D image features with 3D point cloud features for enhanced 3D object recognition. Similarly, BEVGuide\cite{Man2023BEVGuidedMF} employs a cross-attention block using BEV embedding as a guided query to integrate information across different sensors. In our work, we introduce the B2F module to model the relationships between each pixel in the ERP feature map and the entire ICOSAP point feature set, utilizing both semantic-aware affinity attention and distance-aware affinity attention. This enables each ERP pixel feature to capture the spatial and semantic information of large FoV scenes comprehensively. Additionally, our CoCo module employs an attention mechanism to extract spatial context from the depth estimation task and disseminates this information across tasks through spatial attention maps.

\section{Method}
\label{sec:method}
We detail the proposed Elite360M as follows: an \textit{Overview} and \textit{key extensions} over Elite360D in Sec.~\ref{sec:overview}; \textit{Task-Agnostic feature extraction} from ERP and ICOSAP in Sec.~\ref{sec:fea_extract}; the \textit{Bi-projection Bi-attention Fusion (B2F)} in Sec.~\ref{sec:b2f}; the \textit{Cross-task Collaboration (CoCo)} module in Sec.~\ref{sec:coco}; and the \textit{Output heads and Multi-task loss} in Sec.~\ref{sec:loss}.

\subsection{Framework Overview}
\label{sec:overview}

Compared to the depth-focused Elite360D, Elite360M explores the potential of representations obtained through ICOSAP-based bi-projection fusion, simultaneously learning 3D geometry and semantics---thus achieving depth estimation, surface normal estimation, and semantic segmentation within a single training process. Furthermore, it aims to improve the performance of multi-task learning by sharing consistent geometric and semantic information across different tasks. As depicted in Fig.~\ref{fig:overview}, Elite360M comprises three main components: the task-generic feature extraction using ICOSAP-based bi-projection fusion, the Cross-task Collaboration (CoCo) module—designed to segregate task-specific features and integrate spatial contextual information for cross-task fusion—and three task-specific decoders, each committed to generating its respective outputs. Specifically, the encoder includes a flexible ERP feature extractor, an ICOSAP point feature extractor, and a bi-projection bi-attention fusion (B2F) module that learns task-agnostic representations with less distortion and enhanced global perception. Subsequently, the CoCo module employs a scheme of preliminary predictions to disentangle task-specific features and facilitates model interactions among tasks via cross-task attention. Finally, three straightforward decoders with up-sampling and skip connections produce the final predictions. We will provide further details below.

\subsection{Task-Agnostic Feature Extraction}
\label{sec:fea_extract}

\subsubsection{ERP Feature Extraction}
\label{sec:erp_feature}
Taking an ERP image with a resolution of $H \times W$ as input, the encoder extracts a feature map $F^{E} \in \mathbb{R}^{h \times w \times C}$, where $h = H/s$, $w = W/s$, $s$ represents the down-sampling scale factor, and $C$ denotes the number of channels. Our encoder backbone treats ERP images as standard planar images, making it compatible with a broad array of successful networks that have been pre-trained on large-scale planar image datasets, \eg, ImageNet~\cite{JiaDeng2009ImageNetAL}. Notably, the finite size of convolutional kernels or the limited scope of local attention windows means that ERP pixel-wise features lack the necessary global receptive fields for large-FoV content, as local operations cannot replace a global operator~\cite{Wang2017NonlocalNN,Yun2023EGformerEG}. Furthermore, these networks designed for planar images, \eg, ResNet~\cite{He2015DeepRL}, EfficientNet~\cite{Tan2019EfficientNetRM}, Swin transformer~\cite{Liu2021SwinTH},struggle with the inherent spherical distortions present in ERP images, as discussed in Elite360D~\cite{Ai2024Elite360DTE}.
\begin{figure}[!t]
\centering
\includegraphics[width=0.75\linewidth]{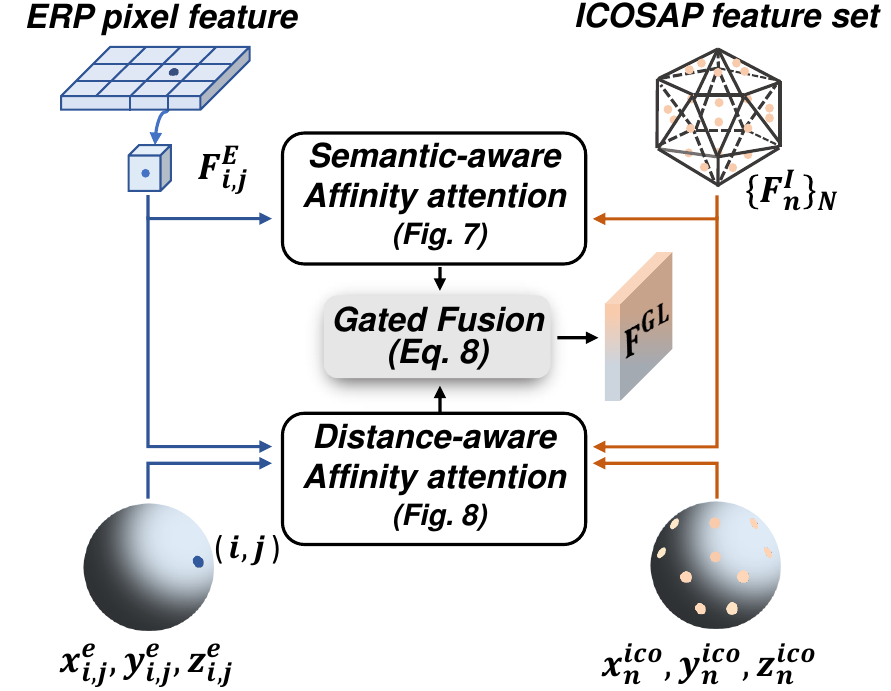}
\vspace{-6pt}
\caption{The overview of B2F, including semantic-aware affinity attention (Fig.~\ref{fig:SAattenion}), distance-aware affinity attention (Fig.~\ref{fig:DAattenion}), and gated fusion (Eq.~\ref{eq:gate_fusion}).}
\label{fig:b2f}
\vspace{-6pt}
\end{figure}
\subsubsection{ICOSAP Feature Extraction} 
\label{sec:ico_feature}
Given that polyhedral representations with a greater number of initial faces exhibit lower distortion~\cite{Lee2018SpherePHDAC}, this paper focuses on the ICOSAP, which has the most initial faces (the initial face number is 20). Especially, we introduce the icosahedron projection (ICOSAP) to provide comprehensive and high-quality global perception. ICOSAP data comprises $20\times4^l$ faces and $12\times4^l$ vertices at the subdivision level $l$ (refer to Fig.~\ref{fig:icosap}). By leveraging the spatial consistency between ICOSAP and ERP within the 3D spherical space, RGB values of ICOSAP vertices can be derived or interpolated from corresponding ERP pixels

Existing works on processing ICOSAP data typically focus on two-dimensional representations, such as unfolded mesh~\cite{Zhang2019OrientationAwareSS} and spherical polyhedron~\cite{Lee2018SpherePHDAC}. These methods involve specifically designed and computationally intensive operations like convolutions, pooling~\cite{Lee2018SpherePHDAC}, and up-sampling~\cite{Shakerinava2021EquivariantNF}. By contrast, we propose to represent ICOSAP data as a point set, as illustrated at the bottom left of Fig.~\ref{fig:overview}. As ERP images already provide dense pixel values, using a discrete ICOSAP point set helps avoid redundancy in semantic information while preserving spatial details and global perception. Specifically, we use the central points of each face to represent the ICOSAP sphere. Initially, we determine the $20\times4^l$ faces of an ICOSAP sphere at the default subdivision level $l$, with each face composed of three vertices. We then calculate the spatial coordinates and RGB values for the center of each face by averaging the values of its three vertices, since each face forms an equilateral triangle. Consequently, we obtain the ICOSAP point set ${P^{ICO}}\in \mathbb{R}^{(20\times4^l)\times6}$, with $20\times4^l$ representing the number of points and $6$ indicating the coordinates [$x, y, z$] and RGB channels. Using the input point set ${P^{ICO}}$, we directly apply the encoder from Point Transformer\cite{Zhao2020PointT} to extract the point feature set ${F^{I}_n}_N\in \mathbb{R}^{N\times C}$, where $N$ is the number of point features and $C$ matches the channel number of the ERP feature map $F^{E}$. For simplicity and efficiency, we do not opt for using complex point-based networks for this purpose.
\begin{figure}[!t]
\centering
\includegraphics[width=0.85\linewidth]{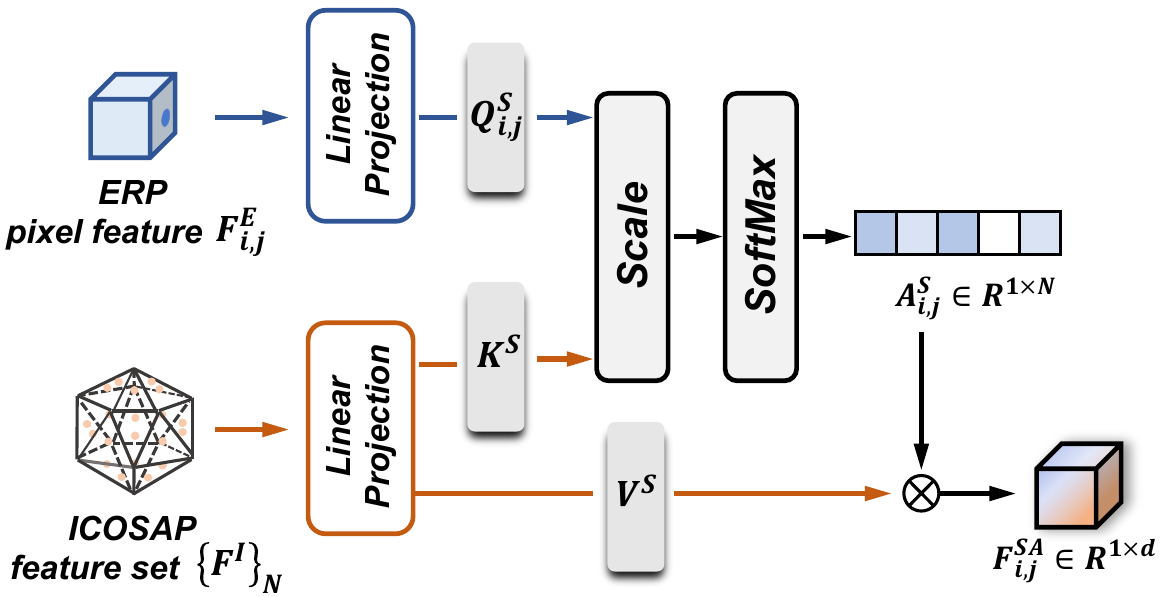}
\vspace{-5pt}
\caption{The architecture of semantic-aware affinity attention. Especially, $\mathbf{Q}^{S}_{i,j} \in \mathbb{R}^{1 \times d}$, $\mathbf{K}^{S} \in \mathbb{R}^{N \times d}$, and $\mathbf{V}^{S} \in \mathbb{R}^{N \times d}$, where $d$ is the dimension and $N$ is the ICOSAP point number.}
\label{fig:SAattenion}
\vspace{-8pt}
\end{figure}

\vspace{-10pt}
\subsection{Bi-Projection Bi-Attention Fusion}
\label{sec:b2f}

In earlier methods such as those described in~\cite{Wang2020BiFuseM3,Wang2022BiFuseSA} and~\cite{Jiang2021UniFuseUF}, bi-projection feature fusion relies heavily on the geometric relationships between CP and ERP. These methods first projects ERP input image into CP patches and obtain the patch-wise CP features. After that, they use the cubemap-to-ERP (C2E) function to back-project CP feature patches into ERP format feature maps, followed by pixel-wise feature concatenation. While this approach is effective, it introduces several issues: 1) Geometry-based re-projection and concatenation-based feature fusion significantly increase computational costs (see Tab.~\ref{tab:com_single_matt}); 2) The geometry-based fusion process from CP patches to ERP panorama limits the global perception of ERP pixels, as each pixel perceives scene information only from its corresponding small-FoV CP patch, without considering other patches; 3) The one-to-one alignment pattern focuses solely on spatial consistency, neglecting semantic similarity. To address these challenges, we have developed the Bi-Projection Bi-Attention Fusion (B2F) module. As illustrated in Fig.~\ref{fig:b2f}, the B2F module first employs semantic-aware affinity attention and distance-aware affinity attention to model the semantic and spatial dependencies between each ERP pixel-wise feature $F^{E}_{i,j}$ and the ICOSAP point feature set ${F^{I}_n}_N$. Here, $(i,j)$ represents the coordinates of a pixel in the ERP feature map, with $i \in (1,h)$ and $j \in (1,w)$, and $N$ denotes the number of points in the ICOSAP feature set. Subsequently, a gated fusion is utilized to adaptively integrate semantic- and distance-related information.

\subsubsection{Semantic-aware affinity attention} 
The semantic-aware affinity attention, as shown in Fig.~\ref{fig:SAattenion}, follows the standard dot-product attention~\cite{dosovitskiy2021an}. Given the extracted ERP feature map $F^{E} \in \mathbb{R}^{h\times w \times C}$ and ICOSAP point feature set $\{F^{I}_n\}_N \in \mathbb{R}^{N \times C}$ (for simplicity, we denote it as $[F^{I}]$), we generate the query $\mathbf{Q}^{S}_{i,j}$ from each ERP pixel-wise feature $F^{E}_{i,j}\in \mathbb{R}^{1 \times C}$ and produce the key $\mathbf{K}^{S}$ and value $\mathbf{V}^{S}$ from the whole ICOSAP point feature set $[F^{I}]$, as
\begin{equation}
\setlength\abovedisplayskip{9pt}
\setlength\belowdisplayskip{9pt}
     \mathbf{Q}^{S}_{i,j} = F^{E}_{i,j}\mathbf{W}^{S}_{Q}, \
     \mathbf{K}^{S} = F^{I}\mathbf{W}^{S}_{K},\mathbf{V}^{S}=F^{I}\mathbf{W}^{S}_{V},
     \label{eq:1}
\end{equation}
where $\mathbf{W}^{S}_{Q}$, $\mathbf{W}^{S}_{K}$, and $\mathbf{W}^{S}_{V}\in\mathbb{R}^{C\times d}$ are linear projections, respectively. $d$ is the output dimension. Then, we calculate the attention weight $\mathbf{A}^{S}_{i,j}$ based on $\mathbf{Q}^{S}_{i,j}$ and $\mathbf{K}^{S}$, and obtain the affinity feature $F^{SA}_{i,j}$:
\begin{equation}
\setlength\abovedisplayskip{9pt}
\setlength\belowdisplayskip{9pt}
     \mathbf{A}^{S}_{i,j} = softmax(\frac{\mathbf{Q}^{S}_{i,j}{\mathbf{K}^{S}}^{T}}{\sqrt{d}}), \ F^{SA}_{i,j} = \mathbf{A}^{S}_{i,j}*\mathbf{V}^{S},
\end{equation}
where $d$ is the channel number and $d=C$. After querying all the ERP pixel-wise features, we can obtain the ERP format feature map $F^{SA}$ with the dimension of $\mathbb{R}^{h\times w\times d}$. The attention weight $\mathbf{A}^{S}_{i,j}$ captures affinities based on the semantic-aware feature similarities, and the output $F^{SA}$ effectively integrates the global and local receptive fields.

\begin{figure}[!t]
\centering
\includegraphics[width=0.90\linewidth]{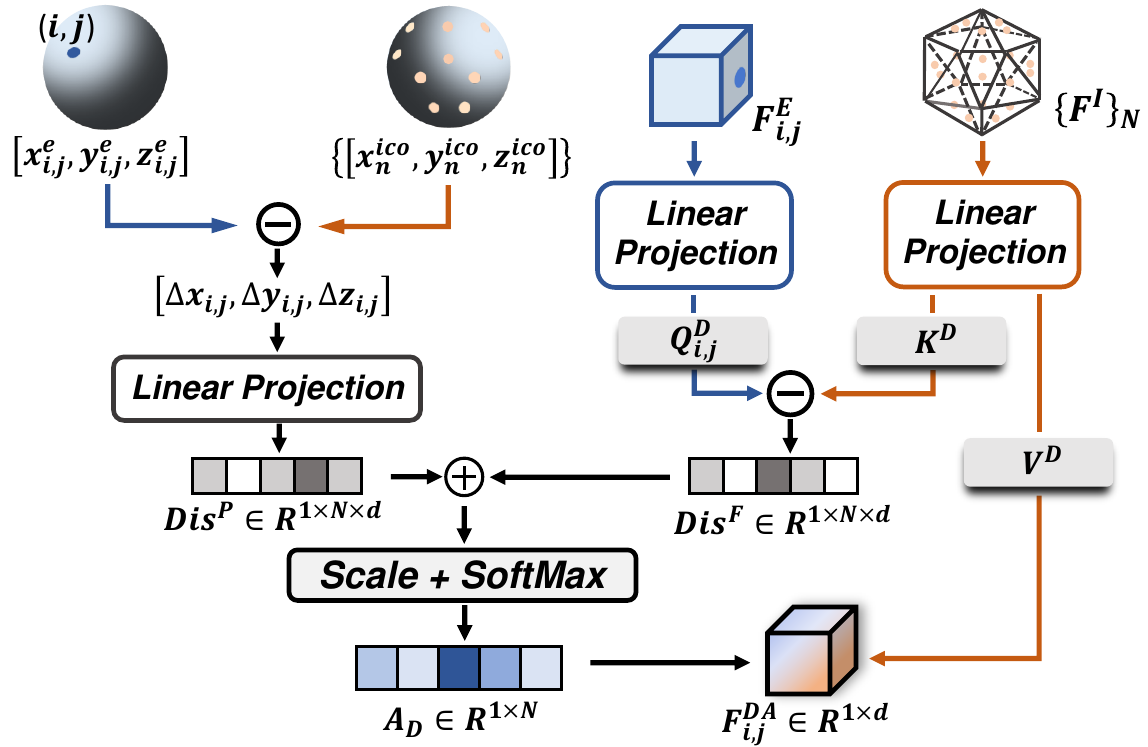}
\vspace{-5pt}
\caption{The architecture of distance-aware affinity attention, especially, $\mathbf{\textit{Q}}^{D}_{i,j} \in \mathbb{\textit{R}}^{1 \times d}$, $\mathbf{\textit{K}}^{D} \in \mathbb{R}^{N \times d}$, and $\mathbf{V}^{D} \in \mathbb{R}^{N \times d}$.}
 \label{fig:DAattenion}
 \vspace{-10pt}
\end{figure}

\begin{figure*}[!t]
\centering
\includegraphics[width=1\linewidth]{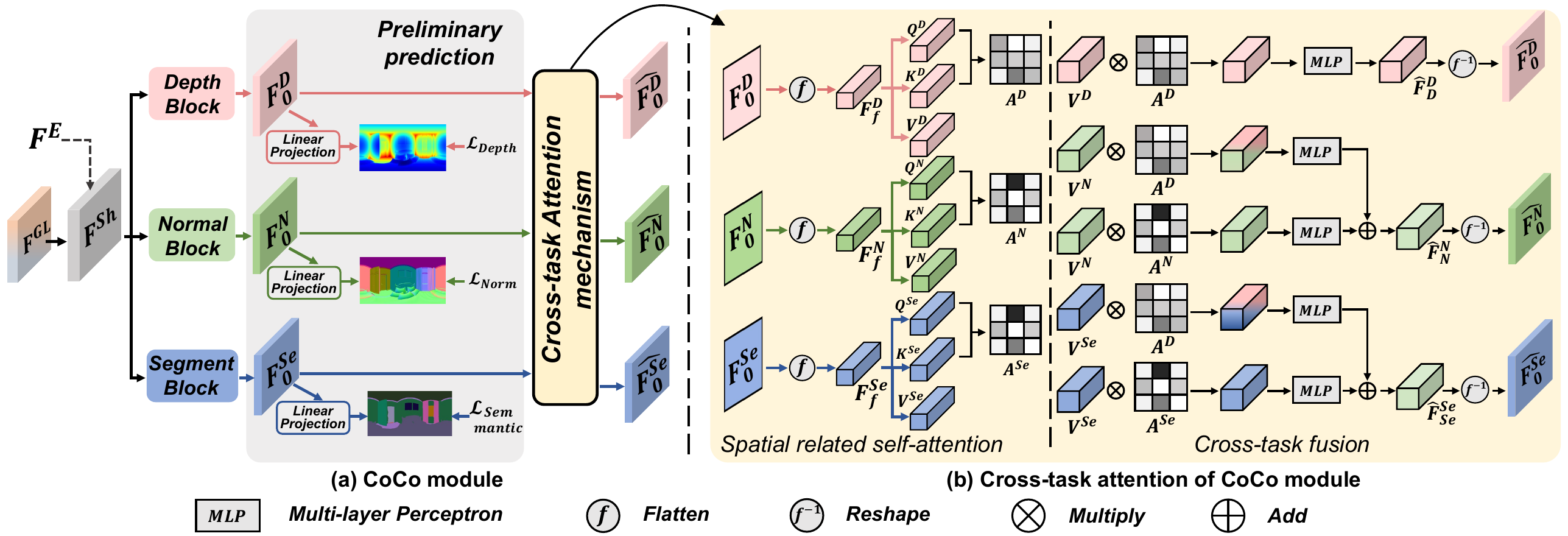}
\caption{(a) The overview of CoCo module, consisting of preliminary predictions and cross-task attention. (b) Structure of the cross-attention mechanism. It first derives spatial attention matrices from task-specific flattened vectors, then facilitates cross-task interactions by sharing the depth-related attention matrix with the other two tasks. Specifically, in the depth estimation branch, the refined feature is produced through self-attention. For the surface normal estimation and semantic segmentation branches, the refined features are generated by combining self-attention features with cross-attention features.}
\label{fig:coco}
\end{figure*}

\subsubsection{Distance-aware affinity attention.} 
Distance-aware affinity attention captures the differences between ERP and ICOSAP in both spatial and semantic information. This can better utilize the geometric prior knowledge of 360$^\circ$ images and precisely measure the distances between two different projection features. As depicted in Fig.~\ref{fig:DAattenion}, the distance-aware affinity attention is built upon the subtraction-based cross-attention mechanism~\cite{wang2022dabert}. Given the ERP pixel-wise feature $F^{E}_{i,j}$ and ICOSAP point feature set $[F^{I}]$, we firstly calculate the spatial distance embedding $Dis^{SP}$ from the spatial coordinates of ERP pixel and ICOSAP point set, 
\ie $[x^{e}_{i,j},y^{e}_{i,j},z^{e}_{i,j}]$ and $\{[x^{ico}_{n},y^{ico}_{n},z^{ico}_{n}]\}_{N}$, as:
\begin{gather}
     Dis^{SP}_{i,j} = [e^{-\varDelta x_{i,j}},e^{-\varDelta y_{i,j}},e^{-\varDelta z_{i,j}}]\mathbf{W}^{SP}, \label{eq:pos} \\ 
     [\varDelta x^{n}_{i,j},\varDelta y^{n}_{i,j},\varDelta z^{n}_{i,j}] = [x^{e}_{i,j},y^{e}_{i,j},z^{e}_{i,j}] - [x^{ico}_{n},y^{ico}_{n},z^{ico}_{n}],\nonumber
\end{gather}
where linear projection $\mathbf{W}^{SP}\in \mathbb{R}^{3\times d}$, $ Dis^{SP}_{i,j} \in \mathbb{R}^{1\times N\times d}$, and $[\varDelta x_{i,j},\varDelta y_{i,j},\varDelta z_{i,j}] \in \mathbb{R}^{1\times N\times 3}$ is the distances between $[x^{e}_{i,j},y^{e}_{i,j},z^{e}_{i,j}]$ and $\{[x^{ico}_{n},y^{ico}_{n},z^{ico}_{n}]\}_{N}$. In particular, the operation $e^{-(\cdot)}$ allows $Dis^{SP}$ to pay more attention to the close parts between ERP pixels and ICOSAP point set. After that, we produce the query $\mathbf{Q}^{D}_{i,j}$ and key $\mathbf{K}^{D}$, and calculate the semantic-related distance embedding $Dis^{SE}_{i,j}$:
\begin{gather}
\vspace{-3pt}
     Dis^{SE}_{i,j} = e^{-\left \| \mathbf{Q}^{D}_{i,j}- \mathbf{K}^{D}\right \|}, \label{eq:dis} \\ 
     \mathbf{Q}^{D}_{i,j}= F^{E}_{i,j}\mathbf{W}^{D}_{Q},\ \ \mathbf{K}^{D}= [F^{I}]\mathbf{W}^{D}_{K},
\vspace{-3pt}
\end{gather}
where $\mathbf{W}^{D}_{Q}$, $\mathbf{W}^{D}_{K}\in \mathbb{R}^{C \times d}$ are linear projections, 
$\mathbf{Q}^{D}_{i,j} \in \mathbb{R}^{1 \times d}$, $\mathbf{K}^{D} \in \mathbb{R}^{N \times d}$, and $Dis^{SE}_{i,j} \in \mathbb{R}^{1\times N \times d}$. Lastly, the distance-aware attention weight $\mathbf{A}^{D}_{i,j}$ is generated with spatial and semantic distance embeddings, and the distance-aware affinity feature vector $F_{i,j}^{DA}$ is obtained from the attention weight $\mathbf{A}^{D}_{i,j}$ 
and the value $\mathbf{V}^{D}$:
\begin{gather}
     \mathbf{A}^{D}_{i,j} = softmax(\frac{{\sum} (Dis^{SP}_{i,j} + Dis^{SE}_{i,j})}{\sqrt{d}}),\\ 
     \mathbf{V}^{D} = F^{I}\mathbf{W}^{D}_{V}, \ \
     F^{DA}_{i,j} = \mathbf{A}^{D}_{i,j}*\mathbf{V}^{D},
\end{gather}
where $\sum $ means the sum for the channel dimension, \ie $\sum(Dis^{SP}_{i,j} + Dis^{SE}_{i,j}) \in \mathbb{R}^{1\times N}$. After querying all ERP pixel-wise features, we obtain the distance-aware aggregated feature $F^{DA}\in\mathbb{R}^{h\times w \times d}$, $d=C$.

\subsubsection{Gated fusion} Since direct average or concatenation may compromise the original representation ability, inspired by~\cite{Cheng2017LocalitySensitiveDN}, we propose the gated fusion block to adaptively fuse $F^{SA}$ and $F^{DA}$ and obtain the representations $F^{GL} \in \mathbb{R}^{h\times w \times C}$ from a local-with-global perspective, formulated as:
\begin{gather}
    F^{GL}= g^{SA} * F^{SA} + g^{DA} * F^{DA},\label{eq:gate_fusion}\\
    g^{SA} = \sigma_{SA} (W^{SA}_g\cdot[F^{SA};F^{DA}]),\nonumber \\
    g^{DA} = \sigma_{DA} (W^{DA}_g\cdot[F^{SA};F^{DA}]),\nonumber
\end{gather}
where $W^{SA}_g$ and $W^{DA}_g$ are linear projections, $\sigma_{SA}(\cdot)$ and $\sigma_{DA}(\cdot)$ are the sigmoid functions.

\subsection{Cross-task Collaboration}
\label{sec:coco}

Inspired by previous methods for planar images~\cite{invpt2022, Ye2023InvPTIP, Xu2022MTFormerML}, we introduce the Cross-task Collaboration (CoCo) module, which includes preliminary predictions and cross-task attention to leverage inter- and cross-task relationships and enhance multi-task learning performance. As depicted in Fig.~\ref{fig:overview}, we first up-sample the task-agnostic representations $F^{GL}$ by a factor of 2 and then establish a skip connection with the corresponding resolution ERP feature map. This process learns a shared representation $F^{Sh} \in \mathbb{R}^{2h \times 2w \times C_1}$, which serves as the input for the CoCo module. Subsequently, within the CoCo module (Fig.~\ref{fig:coco}), preliminary predictions separate task-specific features from the shared representation, and the cross-task attention mechanism transfers attention messages from the depth estimation task to the other two tasks, which can facilitate the sharing of spatial contextual information.

\begin{figure*}[!t]
\centering
\includegraphics[width=\textwidth]{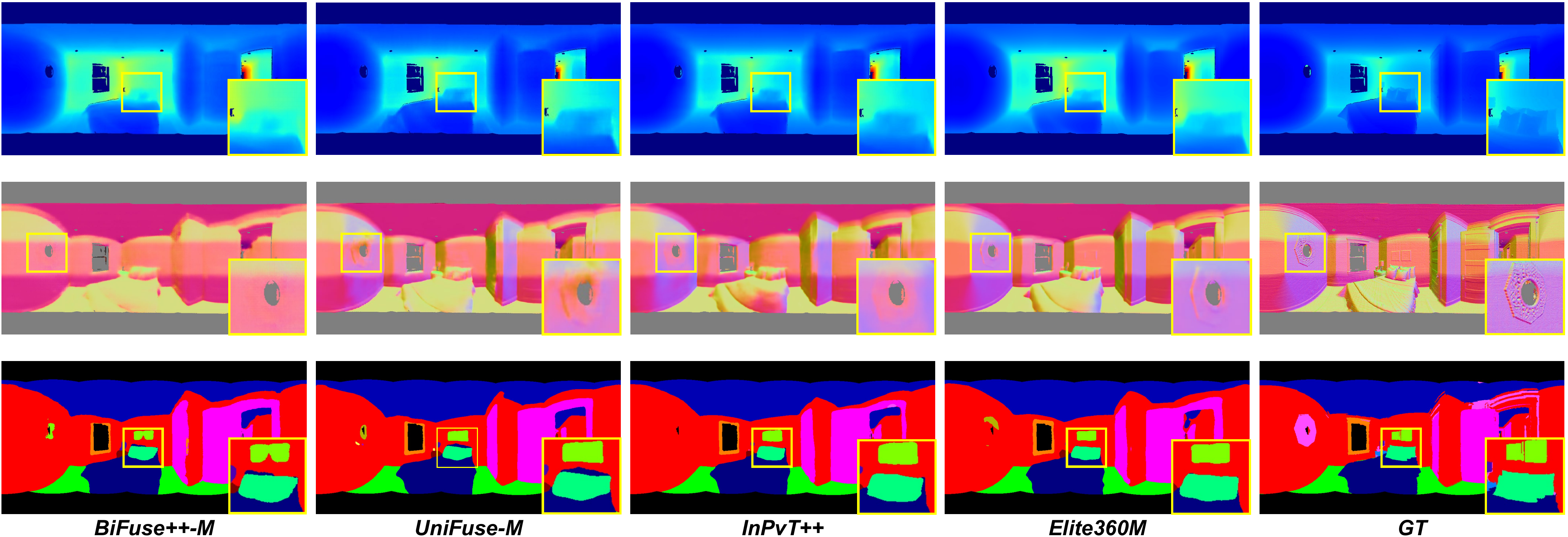}
\vspace{-8pt}
\caption{Qualitative comparison on Matterport dataset among Elite360M and other multi-task learning baselines.}
\vspace{-8pt}
\label{fig:com_multi1}
\end{figure*}
\subsubsection{Preliminary predictions} 
As illustrated in Fig.~\ref{fig:coco}\textred{a}, we initially employ three structurally identical but independent convolutional blocks to extract task-specific features from the shared representation $F^{Sh}$. Each block consists of a convolutional layer with a $3 \times 3$ kernel, followed by batch normalization and ReLU activation. To efficiently separate task-specific geometric and semantic information from the shared representation, we implement the preliminary predictions technique, as described in~\cite{invpt2022,Ye2023InvPTIP}. We introduce three task-specific preliminary output heads to generate initial coarse predictions, which are supervised by down-sampled ground truth labels. This process enables us to extract features rich in task-specific geometric and semantic information, denoted as $F_{0}^{D}$ for depth estimation, $F_{0}^{N} $ for surface normal estimation, and $F_{0}^{Se} \in$ for semantic segmentation, where $F_{0}^{D},F_{0}^{N},F_{0}^{Se} \in \mathbb{R}^{2h \times 2w \times C_1}$. Each preliminary decoder head includes a single linear projection layer. The supervision losses applied are Berhu loss\cite{IroLaina2016DeeperDP} for the depth branch ($\mathcal{L}_{Depth}$), $L1$ loss for the normal branch ($\mathcal{L}_{Normal}$), and classical cross-entropy loss for the segmentation branch ($\mathcal{L}_{Semantic}$).

\subsubsection{Cross-task attention mechanism.} 

After obtaining task-specific features, we introduce a cross-task attention mechanism to model interactions and facilitate effective collaboration among different tasks, as shown in Fig.~\ref{fig:coco}\textred{b}. Unlike traditional message passing and feature fusion between pairs of tasks commonly seen in multi-task learning methods for planar images\cite{Xu2022MTFormerML,vandenhende2020mti}, our approach balances computational costs and performance by leveraging spatial contextual information from one task and sharing it with the other two tasks to achieve efficient cross-task fusion. Building on the success of Elite360D in depth estimation, we generate spatial attention maps from depth features to enrich the normal estimation and semantic segmentation tasks with the geometric and semantic insights from these depth attention maps. Specifically, we first flatten the features from the three tasks to produce $F_{f}^{D}, F_{f}^{N}, F_{f}^{Se}$, and then compute the corresponding self-attention maps $\mathbf{A}^{D}, \mathbf{A}^{N}, \mathbf{A}^{Se}$ as follows:

\begingroup
\footnotesize
\begin{gather}
    \mathbf{Q}^D= F_{f}^{D}\mathbf{W}_{Q}^{D}, \mathbf{K}^D= F_{f}^{D}\mathbf{W}_{K}^{D},\mathbf{A}^{D} = softmax(\frac{\mathbf{Q}^D{\mathbf{K}^D}^T}{\sqrt{d_1}}),\nonumber\\
    \mathbf{Q}^N= F_{f}^{N}\mathbf{W}_{Q}^{N}, \mathbf{K}^N= F_{f}^{N}\mathbf{W}_{K}^{N},\mathbf{A}^{N} = softmax(\frac{\mathbf{Q}^N{\mathbf{K}^N}^T}{\sqrt{d_1}}),\nonumber\\
    \mathbf{Q}^{Se}= F_{f}^{Se}\mathbf{W}_{Q}^{Se}, \mathbf{K}^{Se}= F_{f}^{Se}\mathbf{W}_{K}^{Se},\mathbf{A}^{Se} = softmax(\frac{\mathbf{Q}^{Se}{\mathbf{K}^{Se}}^T}{\sqrt{d_1}}),
\end{gather}
\endgroup
where, $\mathbf{W}_{Q}^{D}, \mathbf{W}_{K}^{D}, \mathbf{W}_{Q}^{N}, \mathbf{W}_{K}^{N}, \mathbf{W}_{Q}^{Se}$, and $\mathbf{W}_{K}^{Se}$ $\in$ $\mathbb{R}^{C_1\times d_1}$ are linear projections. $d_1$ is the channel number and $d_1=C_1$. The self-attention maps $\mathbf{A}^{D}, \mathbf{A}^{N}, \mathbf{A}^{Se}$, defined within $\mathbb{R}^{L \times L}$, capture task-related spatial contextual information, thereby modeling cross-task interactions via self-attention maps. Based on findings that sharing an attention map from one task can enhance the performance of others~\cite{Bhattacharjee2022MuITAE}, we adopt a more efficient approach compared to calculating cross-attention across all three tasks. Given the success of ICOSAP-based bi-projection fusion in depth estimation~\cite{Ai2024Elite360DTE}, we designate the depth estimation task as the primary source for sharing geometric and semantic information. Specifically, for the depth branch, we compute the depth value vector $\mathbf{V}^{D} = F_{f}^{D}\mathbf{W}_{V}^{D}, \mathbf{V}^{D} \in \mathbb{R}^{C_1\times d_1}$ and generate the self-attention output $F^{D}{D} = MLP(\mathbf{A}^{D} * \mathbf{V}^{D})$, where $MLP$ denotes a Multi-Layer Perception layer. After reshaping the resolutions, we obtain the depth feature maps $\widehat{F{0}^{D}} \in \mathbb{R}^{2h\times 2w \times C_1}$, refined by self-attention. For the normal branch, we first compute $\mathbf{V}^{N} = F_{f}^{N}\mathbf{W}_{V}^{N}, \mathbf{V}^{N} \in \mathbb{R}^{C_1\times d_1}$, then calculate the refined normal feature vector $\widehat{F_{N}^{N}} \in \mathbb{R}^{2h\times 2w \times C_1}$ as $\widehat{F_{N}^{N}} = MLP(\mathbf{A}^{N} * \mathbf{V}^{N}) + MLP(\mathbf{A}^{D} * \mathbf{V}^{N})$. Similarly, we derive the semantic feature vector $\widehat{F_{Se}^{Se}}$. Consequently, we can reshape the resolutions to produce the feature maps, denoted as $\widehat{F_{0}^{N}}$ and $\widehat{F_{0}^{Se}} \in \mathbb{R}^{2h\times 2w \times d_1}$.

\begin{table*}[!t]
    \centering
    \caption{Quantitative comparison of multi-task learning frameworks on the Matterport3D test dataset. In terms of model scale, we trained InvPT++ for 100 epochs. \cellcolor{green!20}{Green} indicates the best performance, while \cellcolor{red!20}{Red} denotes the second-best performance. $\dag$ signifies that we re-trained MultiPanoWise under unified settings for a fair comparison.}
    \vspace{-8pt}
    \label{tab:com_multi_matt}
    \resizebox{1\textwidth}{!}{ 
    \begin{tabular}{c|c|c|c|c|c|c|c|c|c|c}
    \toprule[0.8pt]
     \multirow{3}*{Method}&  \multirow{3}*{$\#$Params$^*$ (M)}&\multirow{3}*{$\#$FLOPs$^*$ (G)}& \multicolumn{3}{c|}{Depth} & \multicolumn{3}{c|}{Normal} & \multicolumn{2}{c}{Segmentation}\\
    \cmidrule{4-11}
      & & & Abs Rel $\downarrow$ &$\delta_1 (\%)$ $\uparrow$ &$\delta_2 (\%)$ $\uparrow$& Mean $\downarrow$& $\alpha_{11.25^\circ} (\%) \uparrow$& $\alpha_{22.5^\circ} (\%) \uparrow$&PixAcc \% $\uparrow$&mIoU \% $\uparrow$\\
    \midrule
      InvPT++ (ViT-Large)~\cite{Ye2023InvPTIP}& 379.22&1009.00& \cellcolor{green!20}0.1145 & \cellcolor{green!20}87.68 &\cellcolor{green!20}96.40 & \cellcolor{red!20}18.0715 & 57.00&  76.19 & \cellcolor{green!20}75.28 & \cellcolor{green!20}27.31\\
     \midrule
    UniFuse-M (ResNet-34)~\cite{Jiang2021UniFuseUF} &66.31&158.20&0.1226&86.03&96.23&18.3913&\cellcolor{green!20}57.56&\cellcolor{red!20}76.34&\cellcolor{red!20}74.38&\cellcolor{red!20}26.06\\
    BiFuse++-M (ResNet-34)~\cite{Wang2022BiFuseSA} & 68.13&122.66&0.1344&84.08&95.82&20.4627& 56.41&73.88&73.74&25.37\\
    \midrule
    MultiPanoWise$^\dag$ (PanoFormer)~\cite{shah2024multipanowise} & \textbf{20.51}&148.47& 0.1360& 83.30& 95.29& 20.7391& 56.39 &  73.00& 73.90& 25.52\\
    \midrule
    Elite360M (Ours) (ResNet-34) &29.41&\textbf{85.07}& \cellcolor{red!20}0.1178&\cellcolor{red!20}86.72&\cellcolor{red!20}96.27&\cellcolor{green!20}17.8681&\cellcolor{red!20}57.17&\cellcolor{green!20}76.64&73.42&25.38\\
     \bottomrule[0.8pt]
    \end{tabular}}
    \vspace{-10pt}
\end{table*}
\subsection{Task-specific predictions and Loss functions}
\label{sec:loss}
As depicted in Fig.~\ref{fig:overview}, after refining features via the CoCo module, we feed the refined feature maps of each branch along with the multi-scale feature maps from the ERP decoding backbone into the individual branch decoders. Each branch decoder is equipped with several up-sampling blocks and skip connections. Ultimately, different task-specific output heads generate predictions from the task-specific final-layer features $F^{D}_2$, $F^{N}_2$, and $F^{Se}_2$. Each output head includes a bi-linear up-sampling layer with an up-sampling factor of 4 and a single linear projection layer. The depth output head generates a depth map with a resolution of $H \times W \times 1$. Similarly, the normal output head produces a normal map with a resolution of $H \times W \times 3$, while the segmentation output head yields a segmentation map with dimensions $H \times W \times \text{class}$, where $H \times W$ denotes the resolution of the input ERP image.

For supervision, we do not choose complicated loss functions. Following the approach in~\cite{Kendall2017MultitaskLU}, we weight the loss functions for multiple tasks by considering the homoscedastic uncertainty associated with each task. Specifically, we incorporate three trainable parameters to estimate the uncertainties of the three tasks involved. The formulation of the final loss function is as follows:
\begin{align}
    \mathcal{L}_{totoal} = \mathcal{L}_{Depth} + \mathcal{L}_{Normal}+\mathcal{L}_{Semantic}\nonumber\\
    +\frac{1}{2{\sigma_D}^2} \mathcal{L}_{D} + \frac{1}{2{\sigma_N}^2} \mathcal{L}_{N} + \frac{1}{{\sigma_{Se}}^2} \mathcal{L}_{Se}\nonumber\\
    +\log\sigma_D+\log\sigma_N+\log\sigma_{Se},\label{loss_funcion}
\end{align}
Where $\sigma_D$, $\sigma_N$, and $\sigma_{Se}$ are trainable parameters. $\mathcal{L}_{Depth}$, $\mathcal{L}_{Normal}$, and $\mathcal{L}_{Semantic}$ are the loss functions for preliminary predictions, while $\mathcal{L}_{D}$, $\mathcal{L}_{N}$, and $\mathcal{L}_{Se}$ represent the loss functions for the final predictions in depth estimation, surface normal estimation, and semantic segmentation, respectively. Specifically, we employ the BerHu loss~\cite{IroLaina2016DeeperDP} for depth estimation, the $L1$ loss for surface normal estimation, and the classical cross-entropy loss for semantic segmentation. \textit{More details can be found in the suppl. mat.}.

\section{Experiments}
\label{sec:experiment}
In this section, we present comprehensive experiments and ablation studies to assess the efficacy of Elite360M across two large-scale benchmark datasets. We compare Elite360M's performance with other multi-task learning frameworks (Sec.\ref{sec:experiment_multi}) and single-task learning methods (Sec.\ref{sec:experiment_single}). Additionally, we thoroughly validated and analysed the two core components, B2F and CoCo, in Sec.~\ref{sec:ab_study}.

\subsection{Dataset and Evaluation Metrics}
\label{sec:dataset_metric}

\subsubsection{Dataset}
\label{sec:label}
\begin{figure}[!t]
\centering
\includegraphics[width=1\linewidth]{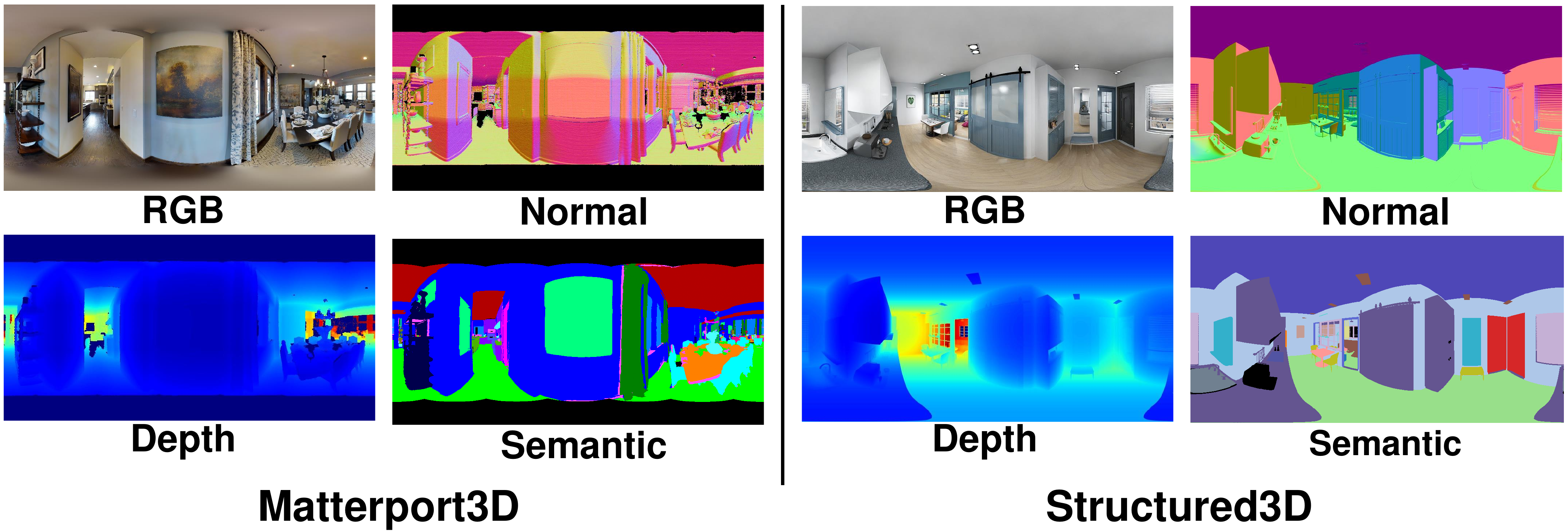}
\vspace{-15pt}
\caption{The samples of two benchmark datasets. \textbf{Left}: real-word Matterport3D; \textbf{Right}: synthetic Structured3D.}
\vspace{-10pt}
\label{fig:dataset}
\end{figure}

We conducted experiments with two benchmark datasets, as depicted in Fig.~\ref{fig:dataset}, each comprising 360$^\circ$ images along with corresponding ground truths for depth, semantic segmentation, and surface normals:
\begin{itemize}
    \item 
    Matterport3D~\cite{Chang2017Matterport3DLF} is a real-world dataset comprising 10,800 360$^\circ$ images from 90 different scenes. Each equirectangular image is stitched together from 18 distinct viewpoints. For depth and surface normal annotations, we converted the provided 18 planar format labels into ERP format. Following the methodology in~\cite{Guttikonda2023SingleFS}, semantic annotations were derived by post-processing the captured point clouds, utilizing open-source software\footnote{\url{https://github.com/atlantis-ar/matterport_utils}}. We adopted the semantic segmentation scheme from~\cite{Chang2017Matterport3DLF}, defining 40 object categories. The dataset is partitioned into training, validation, and test sets according to the protocols specified in~\cite{Chang2017Matterport3DLF}.
    \item 
    Structured3D~\cite{Zheng2019Structured3DAL} is a synthetic scene dataset that comprises 196,515 360$^\circ$ images with diverse illumination and furniture configurations. In this work, we specifically utilize the subset characterized by raw-light illumination and complete furniture arrangements. This subset contains 21,835 samples, for which we employ the standard training, validation, and test splits as outlined in~\cite{Zheng2019Structured3DAL}. For semantic segmentation, following the approach in~\cite{yang2023swin3d}, we select 25 categories that appear with a frequency greater than 0.001 from the original 40 categories.
\end{itemize}

\subsubsection{Evaluation Metrics}
\label{sec:metric}
We employed evaluation metrics from prior work~\cite{Ai2024Elite360DTE, Karakottas2019360SR, Li2023SGAT4PASSSG} as our evaluation criteria. In detail, for the depth estimation task, we measured performance using the Absolute Relative Error (Abs Rel), Root Mean Square Error (RMSE), and a percentage metric with a threshold $\delta_t$, where $t \in {1.25, 1.25^2, 1.25^3}$. For surface normal estimation, we used the Mean Angular Error (Mean) and the percentage of pixel vectors within a preset angle $\alpha_a$, where $a \in {11.25^\circ, 22.5^\circ, 30^\circ}$. The semantic segmentation task was assessed using two common metrics: Mean Intersection over Union (mIoU) and Mean Pixel Accuracy (PixAcc).

\subsection{Implementations of Experiments}
\label{sec:experiment_details}
\subsubsection{Implementation Details}

To ensure a fair comparison, all experiments were conducted under a uniform configuration. For training, we used the Adam optimizer~\cite{Kingma2014AdamAM} with a fixed learning rate of $1 \times 10^{-4}$, consistent with the settings in Elite360D~\cite{Ai2024Elite360DTE}, and we did not use any learning rate scheduler. Furthermore, to rigorously assess the effectiveness of different methods, we retrained each from scratch using their official codes under standardized conditions. The ERP input resolution was set at $512 \times 1024$, and the data augmentation strategies—color augmentation, yaw rotation, and horizontal flipping—were aligned with those used in~\cite{Jiang2021UniFuseUF, Ai2024Elite360DTE}. To ensure convergence, we set the training epochs at 150 for Matterport3D and 60 for Structured3D. All convolution-based backbones, such as ResNet-18 and ResNet-34~\cite{He2015DeepRL}, were pre-trained on ImageNet-1K~\cite{JiaDeng2009ImageNetAL}. Notably, we standardized the default channel numbers for Elite360M to $C = 64$ and $C_1 = 128$. The hyperparameters for the ICOSAP feature extraction were identical to those in Elite360D.
\begin{table*}[!t]
    \centering
    \caption{\textbf{Comparison with the supervised single-task learning methods on Matterport3D test dataset}. $^\dag$ indicates that we adapted these depth estimation methods to achieve surface normal estimation or semantic segmentation tasks. \colorbox{green!20}{Green} represents the best performance using the same backbone, while \colorbox{red!20}{Red} represents the second-best performance.}
    \vspace{-5pt}
    \label{tab:com_single_matt}
    \resizebox{0.85\textwidth}{!}{ 
    \begin{tabular}{c|c|c|c|c|c|c|c|c|c|c|c}
    \toprule[0.8pt]
    \multirow{2}*{Backbone}& \multirow{2}*{Method}& $\#$Params$^*$&$\#$FLOPs$^*$& \multicolumn{3}{c|}{Depth} & \multicolumn{3}{c|}{Normal} & \multicolumn{2}{c}{Segmentation}\\
    \cmidrule{5-12}
     & &(M)&(G)& Abs Rel $\downarrow$ &$\delta_1 (\%)$ $\uparrow$ &$\delta_2 (\%)$ $\uparrow$& Mean $\downarrow$& $\alpha_{11.25^\circ} (\%) \uparrow$& $\alpha_{22.5^\circ} (\%) \uparrow$&PixAcc \% $\uparrow$&mIoU \% $\uparrow$\\
    \midrule[0.8pt]
    Transformer&PanoFormer$^\dag$& 20.38& 81.09 &0.1051 &89.08 &96.23&-&-&-&72.74&24.97\\
    \midrule[0.8pt]
     \multirow{7}*{ResNet-18} & UniFuse$^\dag$~\cite{Jiang2021UniFuseUF} & 30.26 & 62.96 & \cellcolor{green!20}0.1191 & \cellcolor{green!20}86.04& \cellcolor{green!20}95.84 & \cellcolor{green!20}17.4054&\cellcolor{green!20}59.66&  \cellcolor{green!20}78.84&\cellcolor{red!20}72.45 & \cellcolor{green!20}25.35\\
    \cmidrule{2-12}
     & BiFuse++$^\dag$~\cite{Wang2022BiFuseSA} & 32.27 & 54.15 & 0.1387&82.37 &95.03 &18.2394 &57.74&76.29 & 71.66 &23.81\\
    \cmidrule{2-12}
     & HRDFuse$^\dag$~\cite{Ai2023HRDFuseM3} &26.10 & 50.95& 0.1414 & 81.48 & 94.89 & 18.6104 &57.53& 75.97 & 71.20 &23.37\\
    \cmidrule{2-12}
     & Elite360D$^\dag$~\cite{Ai2024Elite360DTE} &15.47&46.27&\cellcolor{red!20}{0.1272} & \cellcolor{red!20}{85.28} & 95.28 & \cellcolor{red!20}{17.8421} &\cellcolor{red!20}{58.64}& \cellcolor{red!20}{76.98} & 71.96 &24.11\\
     &\textbf{Elite360M (Ours)}& \textbf{19.30}& \textbf{65.69} & 0.1340 & 83.64& \cellcolor{red!20}{95.40}&18.5055& 58.00 & 76.85& \cellcolor{green!20}{73.00} & \cellcolor{red!20}{24.50} \\
    \midrule[0.8pt]
     \multirow{7}*{ResNet-34} & UniFuse$^\dag$~\cite{Jiang2021UniFuseUF} &50.48 &96.88 &0.1144 & 87.85 & \cellcolor{green!20}96.59 &\cellcolor{green!20}17.0074& \cellcolor{green!20}60.50 & \cellcolor{green!20}78.43& \cellcolor{green!20}73.88 & \cellcolor{green!20}26.18\\
    \cmidrule{2-12}
     & BiFuse++$^\dag$~\cite{Wang2022BiFuseSA} &52.49 & 87.84& \cellcolor{red!20}0.1123 &\cellcolor{red!20}88.12 &\cellcolor{red!20}96.57 & \cellcolor{red!20}17.2306 &\cellcolor{red!20}59.81& \cellcolor{red!20}77.92 &72.70 &24.03\\
    \cmidrule{2-12}
     & HRDFuse$^\dag$~\cite{Ai2023HRDFuseM3} &46.31& 81.23& 0.1172 & 86.74 & 96.17 & 17.6193 &59.17& 77.39 & 73.07&24.69\\
    \cmidrule{2-12}
     & Elite360D$^\dag$~\cite{Ai2024Elite360DTE} &25.54 & 65.65 & \cellcolor{green!20}0.1115 & \cellcolor{green!20}88.15 & 96.46 & 17.3122 &59.36& 77.86 & 73.14 &25.22\\
     &\textbf{Elite360M (Ours)}&\textbf{29.41} &\textbf{85.07}& 0.1178 & 86.72&96.27 & 17.6880 & 58.17 & 76.64 &\cellcolor{red!20}73.42 &\cellcolor{red!20}25.38\\
     \bottomrule[0.8pt]
    \end{tabular}}
    \vspace{-8pt}
\end{table*}
\begin{table*}[!t]
    \centering
    \caption{Comparison with the supervised single-task learning methods on Structured3D test dataset.}
    \vspace{-2pt}
    \label{tab:com_single_struc}
    \resizebox{0.8\textwidth}{!}{ 
    \begin{tabular}{c|c|c|c|c|c|c|c|c|c}
    \toprule[0.8pt]
    \multirow{3}*{Backbone}& \multirow{3}*{Method}& \multicolumn{3}{c|}{Depth} & \multicolumn{3}{c|}{Normal} & \multicolumn{2}{c}{Segmentation}\\
    \cmidrule{3-10}
     & & Abs Rel $\downarrow$ &$\delta_1 (\%)$ $\uparrow$ &$\delta_2 (\%)$ $\uparrow$& Mean $\downarrow$& $\alpha_{11.25^\circ} (\%) \uparrow$& $\alpha_{22.5^\circ} (\%) \uparrow$&PixAcc \% $\uparrow$&mIoU \% $\uparrow$\\
    \midrule[0.8pt]
     \multirow{6}*{ResNet-34} & UniFuse$^\dag$~\cite{Jiang2021UniFuseUF} &0.1506 & 85.42 & 93.99 &5.4979& 88.88 &92.85& \cellcolor{green!20}94.43 &\cellcolor{green!20}78.09\\
    \cmidrule{2-10}
     & BiFuse++$^\dag$~\cite{Wang2020BiFuseM3} & 0.1666&83.45 &93.46& 5.3412 &88.71&92.71 &94.05 &77.39\\
    \cmidrule{2-10}
     & HRDFuse$^\dag$~\cite{Jiang2021UniFuseUF} & 0.1938 & 78.51& 91.04 & 5.6822 &88.74& 92.54 & 94.03& 77.51\\
    \cmidrule{2-10}
     & Elite360D$^\dag$~\cite{Ai2024Elite360DTE} & \cellcolor{red!20}0.1480 & \cellcolor{green!20}87.41 & \cellcolor{red!20}94.43 & \cellcolor{green!20}4.8246 &\cellcolor{green!20}89.49& \cellcolor{green!20}93.42 & 94.06 & 77.36\\
     &\textbf{Elite360M}& \cellcolor{green!20}0.1430& \cellcolor{red!20}86.71 & \cellcolor{green!20}95.11 & \cellcolor{red!20}4.9131 & \cellcolor{red!20}89.16& \cellcolor{red!20}93.14 & \cellcolor{red!20}94.19 & \cellcolor{red!20}77.78\\
     \bottomrule[0.8pt]
    \end{tabular}}
    \vspace{-10pt}
\end{table*}
\subsubsection{Experimental Settings}
To comprehensively demonstrate the effectiveness of our approach, we devised two sets of comparative settings. In the first set, we selected a range of multi-task learning methods as baselines. These included the state-of-the-art (SOTA) InPvT++\cite{Ye2023InvPTIP}, designed for planar images, the recent MultiPanoWise\cite{shah2024multipanowise}, and frameworks based on UniFuse~\cite{Jiang2021UniFuseUF} and BiFuse++\cite{Wang2022BiFuseSA}. In the second set, we focused on existing single-task learning methods. Specifically, we compared with bi-projection fusion-based methods such as UniFuse, BiFuse++, and HRDFuse\cite{Ai2023HRDFuseM3}. For this comparison, we adapted the final output layers of these depth estimation methods to also support normal estimation and semantic segmentation tasks.

\vspace{-8pt}
\subsection{Performance Comparison}
\label{sec:comparison}
\begin{figure*}[!t]
\centering
\includegraphics[width=0.82\textwidth]{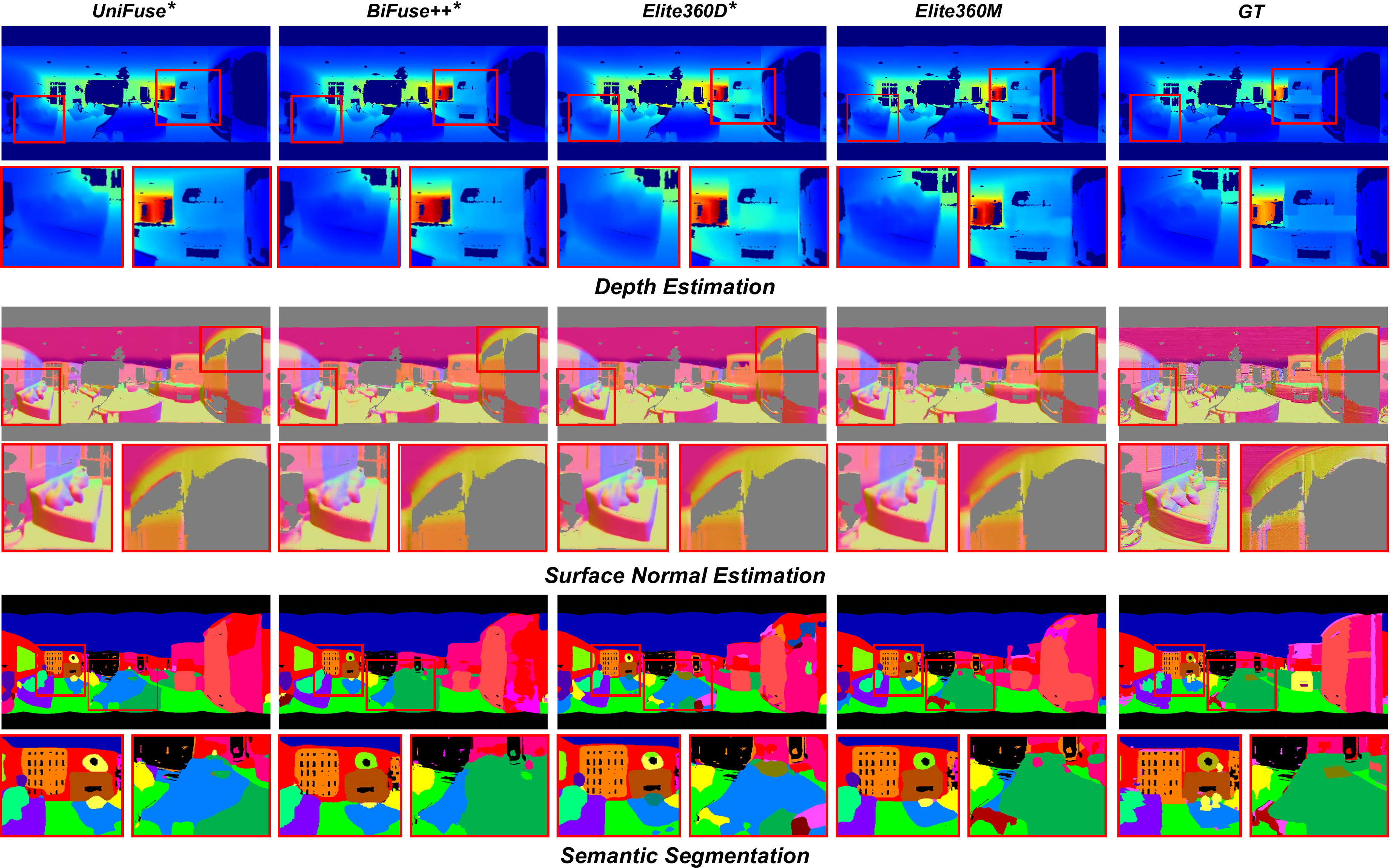}
\vspace{-5pt}
\caption{Qualitative comparison on the Matterport dataset among our multi-task learning method, Elite360M and multiple single-task learning methods. Notably, $*$ denotes that we modify the last decoder layer to adopt the task.}
\vspace{-5pt}
\label{fig:single_com}
\end{figure*}

\subsubsection{Comparison with Multi-task Learning Methods}
\label{sec:experiment_multi}
We conducted a comprehensive comparison on the Matterport3D test dataset with several multi-task learning methods, including InvPT++, MultiPanoWise, and two straightforward multi-task learning baselines derived from UniFuse and BiFuse++. Given that InvPT++ employs the large-scale ViT-Large as its encoder backbone, we trained InvPT++ for 80 epochs on Matterport3D to ensure convergence. We also retrained MultiPanoWise on the Matterport3D training dataset using standardized settings. For the baselines, we adapted the original depth branch structures of BiFuse++ and UniFuse to build three task-specific decoder branches, resulting in two modified frameworks: UniFuse-M and BiFuse++-M. Importantly, for Elite360M, UniFuse-M, and BiFuse++-M, we employed ResNet-34 as the encoder backbone.

The comparative results are displayed in Tab.~\ref{tab:com_multi_matt}, Fig.~\ref{fig:com_multi}, and Fig.~\ref{fig:com_multi1}. Under the integration of the proposed B2F and CoCo modules, Elite360M significantly outperforms bi-projection based baselines in depth estimation and surface normal estimation tasks, achieving a 12.35$\%$ improvement in Abs Rel with BiFuse++-M and a 2.84$\%$ improvement in Mean with UniFuse-M, while using approximately half the parameters and computational resources compared to UniFuse-M and BiFuse++-M. Additionally, as shown in the first row of Tab.~\ref{tab:com_multi_matt}, the computational cost of InPvT++ based on ViT-L exceeds that of our Elite360M by more than tenfold. Despite this, Elite360M outperforms InPvT++ in surface normal estimation and is comparable in depth estimation. Visual results presented in Fig.~\ref{fig:com_multi} and Fig.~\ref{fig:com_multi1} demonstrate that Elite360M predicts more accurate outcomes than bi-projection based baselines and achieves results comparable to the much larger InPvT++ (\textit{Additional visual results are available in the suppl. mat.}). However, Elite360M is less effective in the semantic segmentation task compared to the other two dense pixel regression tasks. One possible reason for this discrepancy in performance gains could be the complex global structural details provided by the B2F module, which may extract fewer semantic details and thereby complicate the classification task. As shown in Fig.~\ref{fig:com_multi}, the prediction results of Elite360M exhibit poor regional smoothness and local consistency.

\begin{table*}[!t]
    \centering
    \caption{\textbf{Quantitative comparison with ERP-based baselines on Matterport3D test dataset}. \cellcolor{gray!40}{\textbf{Bold}} indicates performance improvement.}
    \vspace{-5pt}
    \label{tab:comparison-to-erp}
    \resizebox{1\textwidth}{!}{ 
    \begin{tabular}{c|c|c|c|c|c|c|c|c|c|c|c|c}
    \toprule[1pt]
    \multirow{2}*{Backbone}& \multirow{2}*{Method} &\multirow{2}*{$\#$Params (M)}&\multirow{2}*{$\#$FLOPs (G)}& \multicolumn{4}{c|}{Depth} & \multicolumn{3}{c|}{Normal} & \multicolumn{2}{c}{Segmentation} \\
    \cmidrule{5-13}
    & & & & Abs Rel $\downarrow$ & RMSE $\downarrow$ &$\delta_1 (\%)$ $\uparrow$ &$\delta_2 (\%)$ $\uparrow$& Mean $\downarrow$& $\alpha_{11.25^\circ} (\%) \uparrow$& $\alpha_{22.5^\circ} (\%) \uparrow$&PixAcc \% $\uparrow$&mIoU \%  $\uparrow$\\
    \midrule[1pt]
    \multirow{3}*{Res-18}&Equi-Base& 18.28& 59.63& 0.1525& 0.5597& 80.44& 95.00&18.8167&56.96&76.63& 72.99&24.28\\
    &Ours& 19.30& 65.69& 0.1340 & 0.5395& 83.64 & 95.41&18.5055& 58.00 &76.85&73.00 &24.50\\
    &\cellcolor{gray!40}$\varDelta$ &\cellcolor{gray!40}\textbf{+1.02}& \cellcolor{gray!40}\textbf{+6.06}& \cellcolor{gray!40}\textbf{-12.13$\%$}&\cellcolor{gray!40}\textbf{-3.61$\%$}&\cellcolor{gray!40}\textbf{+3.20$\%$}& \cellcolor{gray!40}\textbf{+0.41$\%$}& \cellcolor{gray!40}\textbf{-1.65$\%$}&
    \cellcolor{gray!40}\textbf{+1.04$\%$}& \cellcolor{gray!40}\textbf{
    +0.22$\%$}&
    \cellcolor{gray!40}\textbf{+0.01$\%$}& \cellcolor{gray!40}\textbf{+0.22$\%$}\\
    \midrule
    \multirow{3}*{Res-34}&Equi-Base& 28.39& 79.01& 0.1242& 0.5174& 85.66& 96.11& 17.9139& 57.44& 76.63& 73.82&25.68\\
    &Ours& 29.41& 85.07& 0.1178 & 0.5047& 86.72 & 96.27&17.8681& 57.18& 76.64&73.42 &25.38\\
    &\cellcolor{gray!40}$\varDelta$ &\cellcolor{gray!40}\textbf{+1.02}& \cellcolor{gray!40}\textbf{+6.06}& \cellcolor{gray!40}\textbf{-5.23$\%$}&\cellcolor{gray!40}\textbf{-2.45$\%$}&\cellcolor{gray!40}\textbf{+1.06$\%$}& \cellcolor{gray!40}\textbf{+0.16$\%$}& \cellcolor{gray!40}\textbf{-0.31$\%$}&
    -0.26$\%$& \cellcolor{gray!40}\textbf{
    +0.01$\%$}&
    -0.40$\%$& -0.30$\%$\\
    \bottomrule[1pt]
    \end{tabular}}
    \vspace{-10pt}
\end{table*}

\subsubsection{Comparison with Single-task Learning Methods}
\label{sec:experiment_single}
For all three tasks, we conducted extensive comparative experiments with supervised single-task learning methods. As demonstrated in Tab.~\ref{tab:com_single_matt} and Tab.~\ref{tab:com_single_struc}, we focused on evaluating our approach against bi-projection-based methods. Therefore, we modified the bi-projection methods designed for depth estimation so that they can accommodate other tasks. Concretely, we tailored only the output channel number of the final layer output heads in these methods to each specific task. Notably, we modified the decoder-level fusion in HRDFuse to feature-level fusion to ensure a more equitable comparison. Our experimental results reveal that our ICOSAP-based B2F module and CoCo module enable Elite360M (\textit{with three decoder branches}) to match or even surpass the performance of bi-projection-based single-task learning methods (\textit{with one decoder branches}), while requiring equivalent or lower computational resources . Specifically, on the Matterport3D dataset with the ResNet-18 backbone (Tab.~\ref{tab:com_single_matt}), Elite360M exceeds BiFuse++ by 3.39$\%$ in Abs Rel and Elite360D by 0.12$\%$ in $\delta_2$ for depth estimation. In surface normal estimation, Elite360M achieves improvements of 0.26$\%$ in $\alpha_{11.25^\circ}$ and 0.56$\%$ in $\alpha_{22.5^\circ}$ over BiFuse++. Remarkably, Elite360M outperforms all single-task learning methods with a 73.00$\%$ PixAcc in semantic segmentation. Moreover, when employing a deeper network, ResNet-34, as the encoder backbone, although the improvements in single-task methods are significant, Elite360M still maintains comparable performance. On the larger-scale synthetic Structured3D, it is evident that for depth estimation, Elite360M surpasses bi-projection fusion-based methods in most metrics, such as a 3.34$\%$ improvement in Abs Rel and a 0.68$\%$ increase in $\delta_2$ compared to the second-best performer, Elite360D. Additionally, Elite360M achieves comparable results to single-task methods in normal estimation and semantic segmentation tasks. (\textit{More qualitative results are shown in the suppl. mat.})

\begin{figure*}[!t]
\centering
\includegraphics[width=1\textwidth]{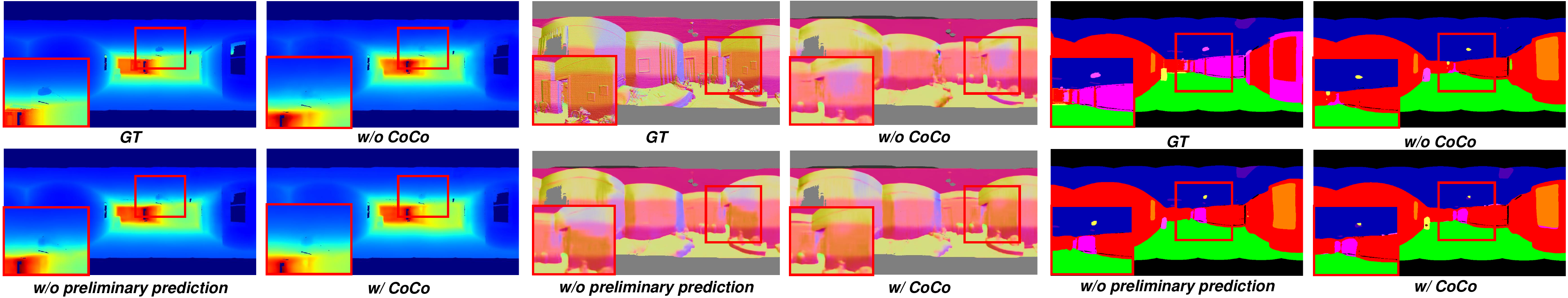}
\vspace{-12pt}
\caption{Visual Comparisons among the components of CoCo module.}
\vspace{-12pt}
\label{fig:CoCo_ab}
\end{figure*}

\vspace{-10pt}
\subsection{Model Analysis}
\label{sec:ab_study}
\subsubsection{Ablation Study of ERP baselines} 
\label{sec:ab_study_erp}

As illustrated in Tab.~\ref{tab:comparison-to-erp}, we compared Elite360M with ERP-based baselines across various ERP encoder backbones on the Matterport3D dataset to assess the capabilities of our proposed bi-projection fusion. With an addition of only approximately 1M parameters, our ICOSAP-based bi-projection fusion achieves substantial improvements across all three tasks. Notably, the performance enhancements are more pronounced when using the lightweight ResNet-18 as the encoder backbone.

Using ResNet-18, Elite360M registers an improvement of over 10$\%$ in Abs Rel error, coupled with a reduction of 1.65$\%$ in Mean error and an increase of 0.22$\%$ in mIoU accuracy. Meanwhile, with the ResNet-34 backbone, our bi-projection fusion significantly enhances the depth estimation task, with a 5.23$\%$ reduction in Abs Rel and a 2.45$\%$ reduction in RMSE. However, performance in normal estimation and segmentation sees a decline with this fusion approach. One potential explanation for this variation is that enhancing the ERP encoder presents substantial challenges for bi-projection fusion in balancing different tasks during multi-task learning.
\begin{table}[t!]
    \centering
     \caption{The ablation results for the CoCo module. We conducted experiments on Matterport3D with the ResNet-34 backbone.}
    \label{tab:ab_coco_part}
    \resizebox{1\linewidth}{!}{ 
    \begin{tabular}{c|c|c|c|c|c|c}
    \toprule[1pt]
      Cross-task &Preliminary& $\#$Params$^*$&$\#$FLOPs$^*$& \textbf{Depth}& \textbf{Normal} & \textbf{Semantic}\\
    attention&prediction& (M)&(G) &$\#$Abs Rel$\downarrow$&$\#$Mean$\downarrow$&$\#$mIoU$\uparrow$ \\
    \midrule[1pt]
     & &28.54 &83.26 & 0.1326 &22.6500 & 25.31\\
    \midrule
   \checkmark & & 29.39& 85.05 & 0.1275 & 21.1093 & 25.34 \\
    \midrule
    \checkmark & \checkmark & 29.41& 85.07& 0.1178&17.6880 & 25.38 \\
    \bottomrule[1pt]
    \end{tabular}}
    \vspace{-5pt}
\end{table}
\subsubsection{Ablation Study of CoCo module}
We performed ablation studies on the Structured3D dataset using ResNet-34 as the ERP encoder backbone to validate the efficacy of the CoCo module. Specifically, we incrementally validated each component within our CoCo module by adding one part at a time. Initially, we emulated the decoder architecture from Elite360D to construct three independent decoder branches for depth, normal, and segmentation. Utilizing the shared representations from the encoder, each decoder outputs its individual predictions. Secondly, We integrated the cross-task attention mechanism to model interactions among the tasks. Finally, we introduced preliminary predictions to more effectively disentangle the task-specific information.

As shown in Table.~\ref{tab:ab_coco_part}, the absence of preliminary predictions and the cross-task attention mechanism significantly reduces the performance of multi-task learning, largely due to the different difficulties of the tasks. Independently predicting each task often complicates achieving convergence for all tasks within a single training process, potentially leading to decreased performance across tasks, as noted in previous studies~\cite{Yun2023AchievementbasedTP, Xu2022MTFormerML}. However, the introduction of the cross-task attention mechanism enhances performance by 3.85$\%$ (Abs Rel) and 6.80$\%$ (Mean). The addition of preliminary predictions further increases performance by 7.61$\%$ (Abs Rel) and 16.21$\%$ (Mean). This comparison shows that preliminary predictions effectively extract task-specific information from shared representations, thereby enhancing multi-task learning accuracy. However, the improvement in semantic segmentation from the CoCo module is less pronounced compared to depth and normal estimation. This may be due to the relatively simple structural information in semantic labels, which limits the impact of cross-task attention. We also present qualitative comparisons in Fig.~\ref{fig:CoCo_ab}, which visually illustrate the influence of each component more effectively.
\begin{table}[!t]
    \centering
    \vspace{-10pt}
    \caption{The ablation results for B2F module for depth estimation.}
    \label{tab:ab_biprojection_fusion}
    \resizebox{1\linewidth}{!}{ 
    \begin{tabular}{c|c|c|c|c}
    \toprule
    Bi-projection feature fusion & Abs Rel $\downarrow$&Sq Rel $\downarrow$&RMSE $\downarrow$& $\delta_1$ $\uparrow$\\
    \midrule
    SFA~\cite{Ai2023HRDFuseM3} + Add & 0.1276& 0.1002 &0.5150 & 84.27\\
    SFA~\cite{Ai2023HRDFuseM3} + Concat & 0.1191& 0.1019 & 0.5143& 86.52\\
    Only SA & 0.1204& 0.1014 &  0.5121& 86.26\\
    Only DA &  0.1184 & 0.0972 & 0.4944 &87.06 \\
    Our B2F (SA + DA) & \textbf{0.1115} & \textbf{0.0914} & \textbf{0.4875} &  \textbf{88.15} \\
    \bottomrule
    \end{tabular}}
    \vspace{-10pt}
\end{table}

\subsubsection{Ablation Study of B2F module}
For the ablation analysis of the B2F module, we primarily focus on its performance in the depth estimation task.
\begin{itemize}
    \item \noindent \textit{The effectiveness of B2F module.} In Tab.~\ref{tab:ab_biprojection_fusion}, we compared four available bi-projection feature fusion modules. To align the spatial dimensions between ICOSAP point feature set and ERP feature map, we introduce SFA module 
    from~\cite{Ai2023HRDFuseM3}. After that, we employed direct addition and concatenation to aggregate these two projections. We also achieved the bi-projection feature fusion with semantic-aware affinity attention (SA) alone and distance-aware affinity attention (DA) alone. Compared to the methods based solely on semantic-aware feature similarities (The first three rows), single distance-aware affinity attention can achieve better performance, which indicates that the spatial positional relationships boost the bi-projection feature fusion. Overall, our B2F outperforms others.
    \item \noindent \textit{The superiority of ICOSAP.} As only CP$/$TP's patch centers lie on the sphere's surface, we extracted the feature embedding from each CP$/$TP patch and employs the patch center coordinates and feature embedding as the input of B2F module. In Tab.~\ref{tab:ab_diff_proj}, we show the results with ResNet18 backbone. Our bi-projection fusion procedure, utilizing the ICOSAP point set, marginally outperforms the fusion between CP and TP patches, while exhibiting fewer parameters and FLOPs.
\end{itemize}
\begin{table}[!t]
\centering
    \caption{The comparison of different projections on Matterport3D for depth estimation task.}
    \label{tab:ab_diff_proj}
    \vspace{-5pt}
\resizebox{1\linewidth}{!}{ 
    \begin{tabular}{cccccc}
    \toprule
    Method&$\#$Param(M) &$\#$FLOPs(G) &Abs Rel $\downarrow$ & RMSE $\downarrow$  &$\delta_1$ $\uparrow$ \\
    \cmidrule{1-6}
   ERP-CP& 25.66 &  54.15 &  0.1369& 0.5401& 83.69\\
   \cmidrule{1-6}
   ERP-TP (N=18) & 25.66& 50.58 & 0.1328& 0.5385& 83.87 \\
    \cmidrule{1-6}
    ERP-ICOSAP (Ours) &\textbf{15.43}&\textbf{45.91} &\textbf{0.1272} & \textbf{0.5270} &\textbf{85.28}\\
    \bottomrule
    \end{tabular}}
    \vspace{-10pt}
\end{table}
\subsection{Discussions}

Experimental results presented in Tab.~\ref{tab:com_single_matt} reveal that there is still a gap between our approach and single-task learning methods on real-world scenarios. However, our Elite360M can simultaneously perform multiple perception tasks using fewer model parameters and computational resources. This efficiency makes Elite360M suitable for deployment on edge devices, offering practical value for applications like robotic navigation and autonomous driving, which require concurrent perception of geometric and semantic information for comprehensive scene understanding. Furthermore, as indicated by the results in Tab.~\ref{tab:com_multi_matt} and Tab.~\ref{tab:comparison-to-erp}, the performance in the semantic segmentation task is not as robust as in the other two tasks. We attribute this disparity primarily to the integration of ICOSAP. Given that our implementation utilizes a simple, untrained point transformer to extract ICOSAP features, these features are more geometrically than semantically informative, slightly undermining performance in semantic segmentation tasks.

\section{Conclusion}
\label{sec:conclusion}
In this paper, we introduced Elite360M, a novel multi-task learning framework that learns 3D geometry and semantics from 360$^\circ$ images to simultaneously accomplish three scene understanding tasks. Compared with existing bi-projection fusion schemes based on CP patches or TP patches, we introduced non-Euclidean ICOSAP and presented a compact yet effective Bi-projection Bi-attention Fusion module to learn a less-distorted and more globally perceptive representation from the combination of ERP and ICOSAP. Furthermore, we introduced the Cross-task Collaboration module to boost multi-task learning performance with two steps, which firstly leverages preliminary predictions to extract task-specific features from the task-agnostic representation and then share spatial contextual messages via attention maps to achieve cross-task fusion. Through extensive experiments, we validated the efficiency and effectiveness of each component. Furthermore, with the cooperation of our proposed modules, our Elite360M outperforms the multi-task learning baselines and is on par with single-task learning methods. In the future, we will explore the development of specialized encoders for icosahedron projection to better learn semantics from this non-Euclidean projection format data and achieve the performance improvement on semantic segmentation. Indisputably, it would be highly beneficial for the 360 vision community to provide a specific network, which could balance the geometry and semantics of ICOSAP point set.
{
\bibliographystyle{IEEEtran}
 \bibliography{egbib}
}

\begin{IEEEbiography} [{\includegraphics[width=1in,height=1.2in,clip,keepaspectratio]{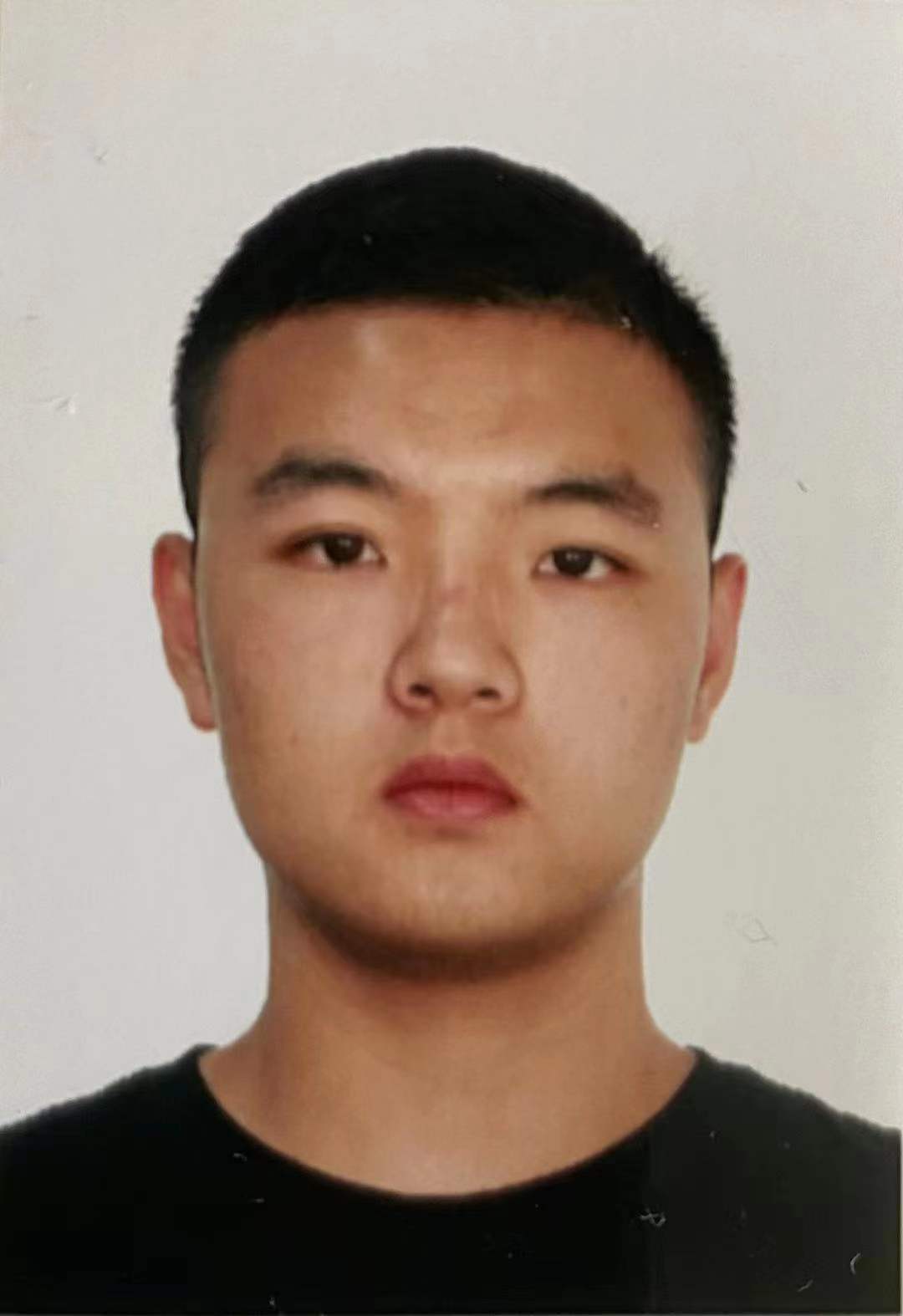}}] {Hao Ai}
is a Ph.D. student in the Visual Learning and Intelligent Systems Lab,  Artificial Intelligence
Thrust,  Guangzhou Campus, The Hong Kong University of Science and Technology (HKUST). His research interests include pattern recognition (image classification, face recognition, etc.), DL (especially uncertainty learning, attention, transfer learning, semi- /self-unsupervised learning), omnidirectional vision.
\end{IEEEbiography}
\begin{IEEEbiography}[{\includegraphics[width=1in,height=1.2in,clip,keepaspectratio]{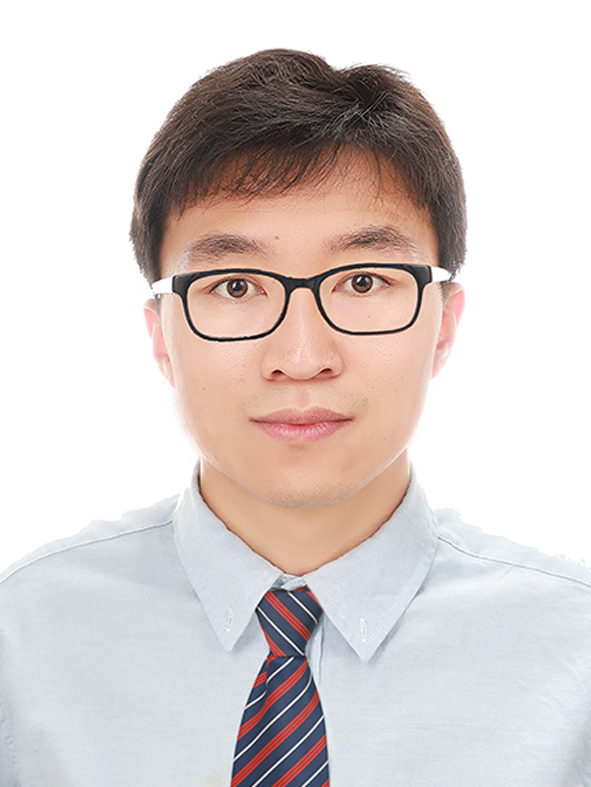}}]{Lin Wang} is an assistant professor in the Artificial Intelligence Thrust, HKUST GZ Campus, HKUST Fok Ying Tung Research Institute, and an affiliate assistant professor in the Dept. of CSE, HKUST, CWB Campus. He got his PhD degree with honor from Korea Advanced Institute of Science and Technology (KAIST).  His research interests include computer vision/graphics, machine learning and human-AI collaboration.
\end{IEEEbiography}
\vfill

\end{document}


\title{Elite360M: Efficient 360 Multi-task Learning via Bi-projection Fusion and Cross-task Collaboration \\--- Supplemental Material ---}

\maketitle
\begin{abstract}
Due to the lack of space in the main paper, we provide more details of the proposed Elite360M and experimental results in the supplementary material. In Sec.~\ref{sec:supp_icosap}, we provide more prior knowledge of icosahedron projection (ICOSAP). In Sec.~\ref{sec:supp_network}, we provide a more detailed illustration of our network architecture. Sec.~\ref{sec:supp_loss} introduces our training loss functions and Sec.~\ref{sec:supp_dataset} introduces the used datasets and evaluation metrics in details. Moreover, we introduce the specific training details of prevalent methods and ERP-based in Sec.~\ref{sec:supp_training_detail}, and we show additional quantitative and qualitative comparison results in Sec.~\ref{sec:supp_comparison_results}.

\end{abstract}
\section{Icosahedron Projection}
\label{sec:supp_icosap}
\begin{figure}[ht]
    \centering
    \resizebox{1\linewidth}{!}{ \includegraphics{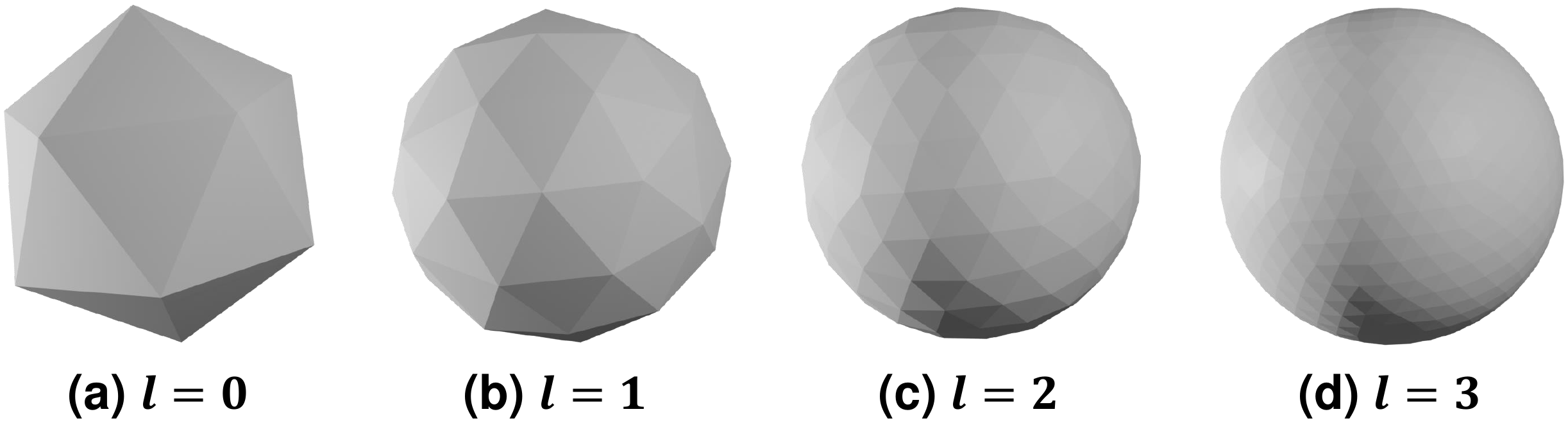}}
    \caption{The subdivision of icosahedron at different resolution $l$.}
    \label{fig:supp_ico}
\end{figure}

\begin{figure}[ht]
    \centering
    \resizebox{1\linewidth}{!}{ \includegraphics{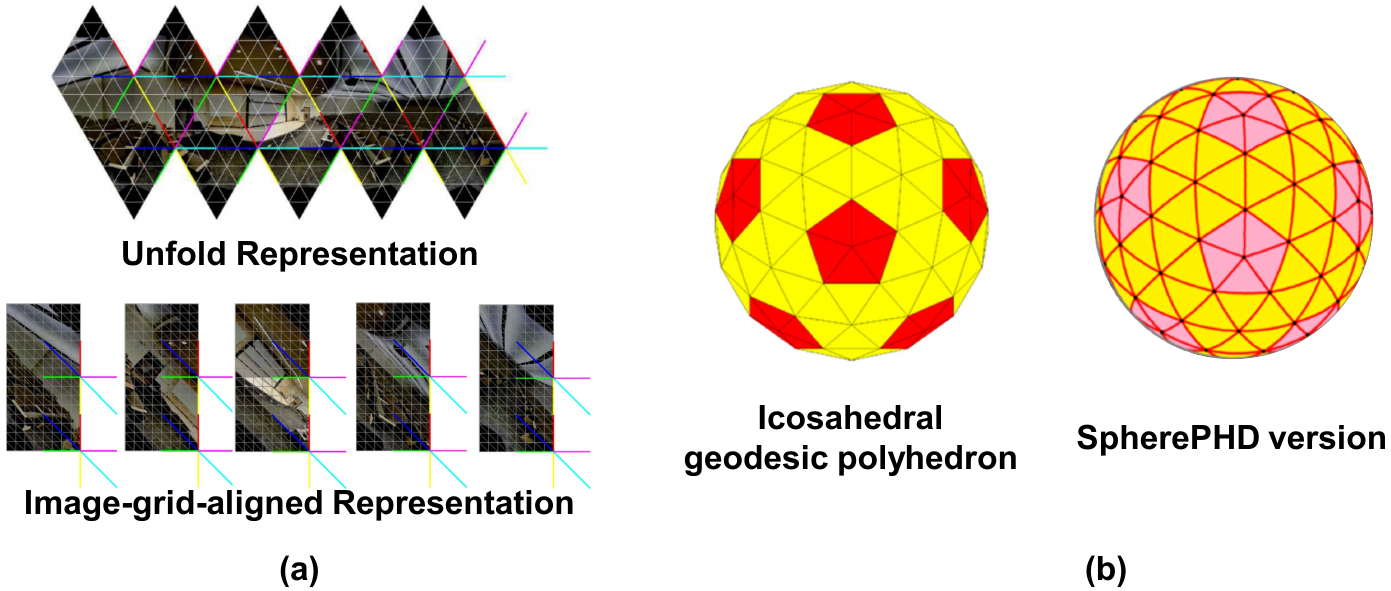}}
    \caption{(a) Unfold representation (top) and image-grid-aligned representation (bottom) of ICOSAP. (b) icosahedral spherical polyhedron (SpherePHD) representation. (a), (b)are originally shown in~\cite{Zhang2019OrientationAwareSS} and~\cite{Lee2018SpherePHDAC}, respectively.}
    \label{fig:supp_ico_representation}
\end{figure}

As shown in Fig.~\ref{fig:supp_ico}, with each increment in resolution on icosahedron projection (ICOSAP), each subdivision divides each equilateral triangle into 4 smaller equilateral triangles. We designate the initial icosahedron projection as $l=0$, where $l$ represents the level of subdivision. For different subdivision levels of the icosahedron projection, we can obtain corresponding faces ($f=20\times 4^{l}$) and vertices ($v=2+ 1\times 4^{l}$), as shown in Tab.~\ref{tab:sup_ico}. In the existing methods~\cite{Lee2018SpherePHDAC,Zhang2019OrientationAwareSS,Shakerinava2021EquivariantNF}, they focus on enable operations, \eg, convolution, pooling, and padding, on the ICOSAP. Specifically,~\cite{Zhang2019OrientationAwareSS} unfolds the ICOSAP onto the plane and aligns it to the standard image grid (See Fig.~\ref{fig:supp_ico_representation}(a)), while~\cite{Lee2018SpherePHDAC} represents 360$^\circ$ image as a SpherePHD version of geodesic icosahedron and directly presents the sphere convolution layers and pooling layers to process it (See Fig.~\ref{fig:supp_ico_representation}(b)). 
\begin{table}[ht]
    \centering
        \caption{The subdivision of ICOSAP. The red points are the vertices and the original points are face centers.}
    \label{tab:sup_ico}
    \resizebox{0.8\linewidth}{!}{ 
    \begin{tabular}{c|c|c|c}
    \toprule
     Resolution& $\#$Faces &$\#$Vertices&Each face\\
    \midrule
        $l=0$& 20 & 12 & \begin{minipage}[b]{0.3\columnwidth}
		\centering
		\raisebox{-.5\height}{\includegraphics[width=\linewidth]{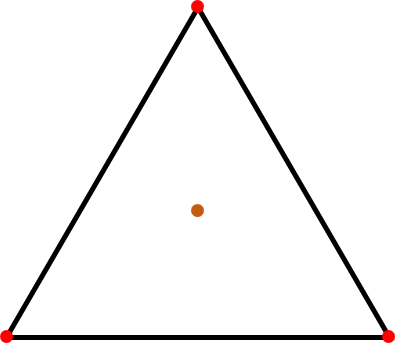}}
	\end{minipage}  \\
    \midrule
        $l=1$& 80 & 42 & \begin{minipage}[b]{0.3\columnwidth}
		\centering
		\raisebox{-.5\height}{\includegraphics[width=\linewidth]{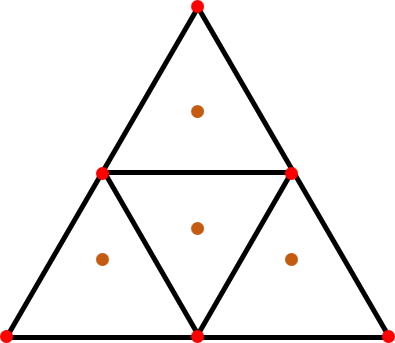}}
	\end{minipage}  \\
    \midrule
        $l=2$& 320 & 162 & \begin{minipage}[b]{0.3\columnwidth}
		\centering
		\raisebox{-.5\height}{\includegraphics[width=\linewidth]{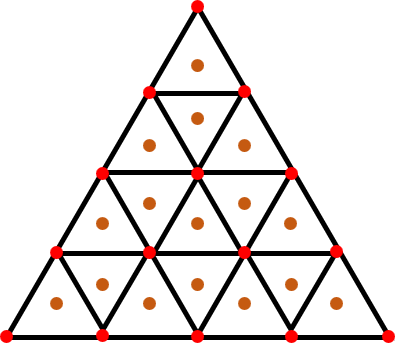}}
	\end{minipage}  \\
        \midrule
        $l=n$& $20\times 4^{n}$ & $2+10\times 4^{n}$ & $\cdots$ \\
    \bottomrule
    \end{tabular}}
\end{table}

\begin{figure}[ht]
    \centering
    \resizebox{0.8\linewidth}{!}{ \includegraphics{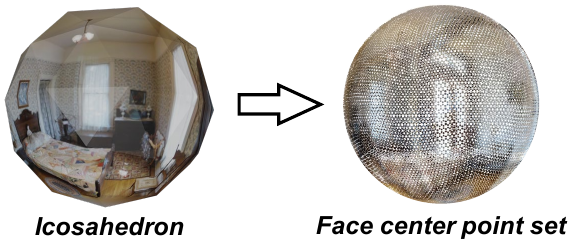}}
    \caption{ICOSAP and corresponding face center point set.}
    \label{fig:supp_ico_center}
\end{figure}
\begin{figure*}[t]
    \centering
    \resizebox{1\textwidth}{!}{ \includegraphics{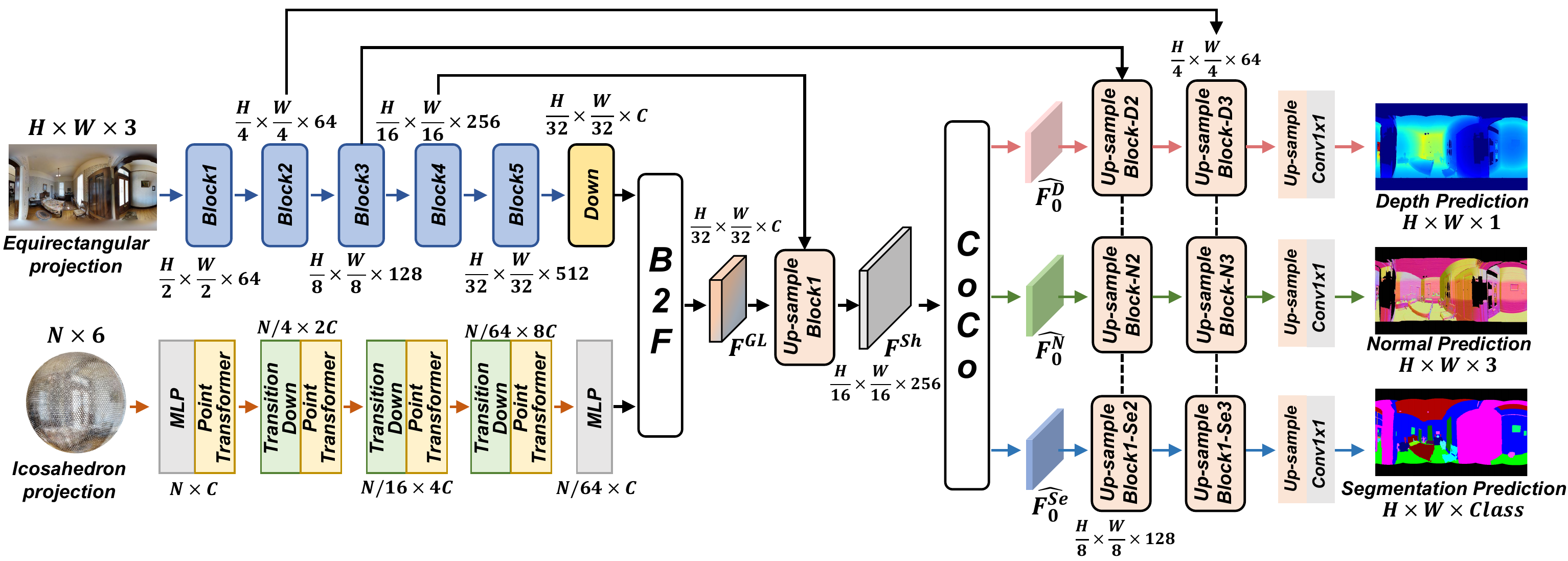}}
    \caption{Overall Architecture of Elite360M.}
    \label{fig:supp_nework_achitecture}
\end{figure*}
\section{Details of Network Architecture}
\label{sec:supp_network}
In our Elite360M, for efficiency, we represent the ICOSAP data as the discrete point set. For better comprehension, we use a subdivision level $l=4$ as an example. Firstly, we create the subdivided icosahedron
with the $20 \times 4^{4} =5120 $ faces and $2+10 \times 4^4 = 2562$ vertices. Especially, as the vertices of the icosahedron lie on a spherical surface, we can calculate the RGB values of these vertices using the spherical spatial correspondence between the vertices and equirectangular projection (ERP) pixels. After obtaining the spatial coordinates $\left(x, y, z\right)$ of the vertices along with their RGB values, we can interpolate to obtain the coordinates and RGB values of the face centers. Since each face of the icosahedron is an equilateral triangle, we can directly calculate the averages of vertices for faces' coordinates and RGB values. Finally, we collect the $20 \times 4^{4} =5120$ face centers (See Fig.~\ref{fig:supp_ico_center}) and obtain the ICOSAP input tensors with the dimension of $\mathbb{R}^{5120 \times 6}$.

\begin{figure}[ht]
    \centering
    \resizebox{0.65\linewidth}{!}{ \includegraphics{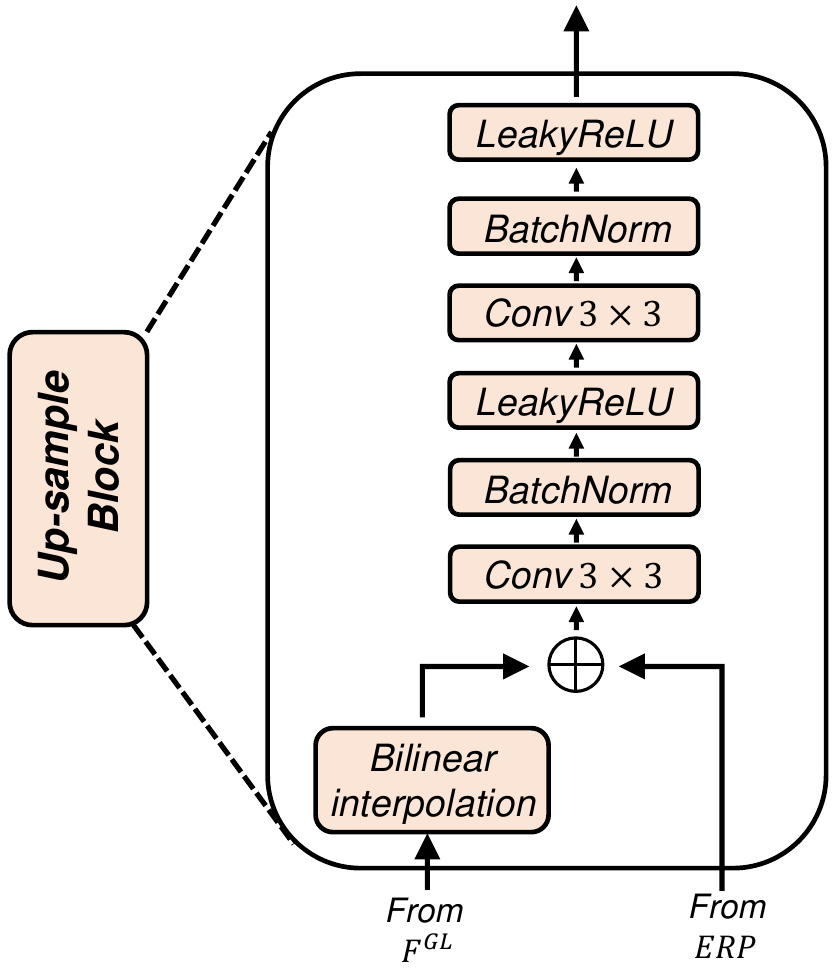}}
    \caption{Up-sample block in the decoder. $\bigoplus$ means the concatenation.}
    \label{fig:supp_nework_upsample}
\end{figure}

The overall network structure is described in Fig.~\ref{fig:supp_nework_achitecture}. The network comprises five main components: ERP feature extraction, ICOSAP feature extraction, B2F module, CoCo module and task-specific decoder heads. In most experimental setups, both ERP feature extraction and ICOSAP feature extraction have a channel number $C$ of 64. Similarly, the B2F module also operates with $d=64$ channels. For the channel number $C_1$ in the CoCo module, we set it as 256. Especially, for ERP feature extraction, we directly employ the 2D models, which are designed for perspective images and pre-trained on large-scale ImageNet~\cite{JiaDeng2009ImageNetAL} dataset. For ICOSAP feature extraction, we directly employ the encoder in Point Transformer~\cite{Zhao2020PointT} with three transition down blocks as the point feature extractor. Especially, for the decoding stage, we first up-sample the fused feature $F^{GL}$ by two times with the provided up-sampling block and then concatenate the up-sampled fused feature with corresponding ERP feature map via the common skip-connection. Subsequently, with the obtained shared representation $F^{Sh}$, we employ the CoCo module to disentangle task-specific information from $F^{Sh}$ and model the cross-task interactions to boost multi-task learning. After learning three representations, $\widehat{F_{0}^{D}}$, $\widehat{F_{0}^{N}}$, and $\widehat{F_{0}^{Se}}$, we repeat the prior procedure, utilizing up-sampling blocks to double the resolution sizes of the feature maps and employing skip connections to integrate the features of each task with the ERP features. Finally, through output heads, we generate predictions for the three tasks. Each output heads consist of a up-sampling layer with the up-sampling factor of 4 and The specifics of the up-sampling block are illustrated in Fig.~\ref{fig:supp_nework_upsample}.

\section{Training Loss}
\label{sec:supp_loss}
For the entire training process, our objective function consists of two parts: the first part is the loss functions for preliminary predictions, while the second part is derived from the loss functions for the final outputs. As for preliminary predictions, we simply sum the losses from the three tasks and set the weight of each loss to 1. Specifically, we employ Berhu loss~\cite{IroLaina2016DeeperDP} for the depth supervision, L1 loss for the surface normal supervision, and classical entropy loss for semantic segmentation supervision. For the final outputs, we follow existing works~\cite{Xu2022MTFormerML,Kendall2017MultitaskLU} and employ the uncertainty loss, which weighs multiple loss functions by considering the uncertainty of each task. Specifically, for each task, the loss function is consistent with that used in the preliminary predictions. In total, the overall loss function can be written as:
\begin{equation}
\small
\begin{gathered}
\mathcal{L}_{totoal} = \mathcal{L}_{pre} + \mathcal{L}_{final},\\ 
\mathcal{L}_{pre}= \mathcal{L}_{D}(D_{pre},D^{GT}_{pre}) + \mathcal{L}_{N}(N_{pre},N^{GT}_{pre})+ \mathcal{L}_{Se}(Se_{pre},Se^{GT}_{pre}),\\
\mathcal{L}_{final} = \frac{1}{2{\sigma_D}^2} \mathcal{L}_{D}(D,D^{GT}) + \frac{1}{2{\sigma_N}^2} \mathcal{L}_{N}(N,N^{GT}) \\ 
+ \frac{1}{{\sigma_{Se}}^2} \mathcal{L}_{Se}(Se,Se^{GT})+\log\sigma_D+\log\sigma_N+\log\sigma_{Se},\\
\end{gathered}
\label{eq:loss_function}
\end{equation}
where $D_{pre}$, $N_{pre}$, and $Se_{pre}$ represent the preliminary predictions, and $D^{GT}{pre}$, $N^{GT}{pre}$, and $Se^{GT}_{pre}$ are the corresponding ground truths, matched in scale to the preliminary predictions through the down-sampling operation. $D$, $N$, and $Se$ are the final outputs, and $D^{GT}$, $N^{GT}$, and $Se^{GT}$ are the ground truths. $\sigma_D$, $\sigma_N$, and $\sigma_{Se}$ are trainable values to estimate the uncertainties. The task-specific loss functions can be written as:
\begin{equation}
\small
\begin{gathered}
\mathcal{L}_{D}= \frac{1}{N} \sum^{N}_{i=1} \mathcal{B}(x^i,GT^{i}), \\ 
\mathcal{B}(x^i,GT^{i}) = \left\{\begin{aligned}	
		&\left | x-GT \right |, \left | x-GT\right |\leq c\\
		&\frac{(x-GT)^2+c^2}{2c}, \left | x-GT\right | > c\\
	\end{aligned}\right. \\ 
\mathcal{L}_{N}(x,GT) = \frac{1}{N} \sum^{N}_{i=1}\left | x^i-GT^{i}\right | \\
\mathcal{L}_{Se}(x,GT) = \frac{1}{N} \sum^{N}_{i=1}CE(x^i, GT^i)
\end{gathered}
\label{eq:task_loss}
\end{equation}
where $x$ represents the predicted results, GT represents the ground truths, and N is the batch size. $\mathcal{B}$ is the Berhu loss, and $c$ is a threshold hyper-parameter and set to 0.2 empirically~\cite{Li2022OmniFusion3M,Jiang2021UniFuseUF}.

\begin{table}[t]
    \centering
        \caption{The statistics of the datasets.}
    \label{tab:sup_dataset}
    \resizebox{0.80\linewidth}{!}{ 
    \begin{tabular}{c|c|c|c}
    \toprule
    Datasets&$\#$train&$\#$val&$\#$test\\
    \midrule
    Matterport3D& 7829&957&2014\\
    Structured3D&18343 &1776&1697\\
    \bottomrule
    \end{tabular}}
\end{table}

\begin{figure*}[t]
    \centering
    \resizebox{0.8\textwidth}{!}{ \includegraphics{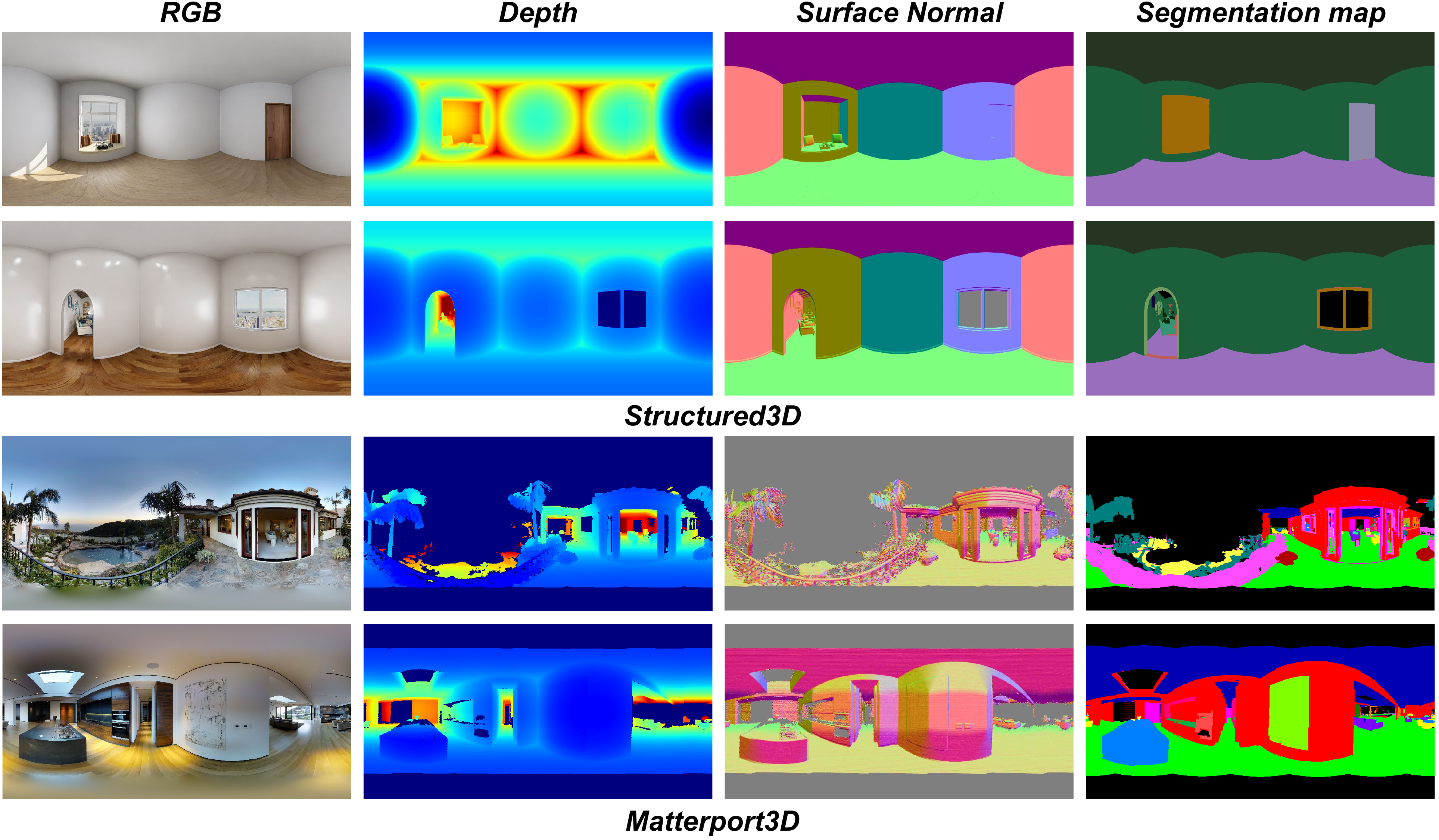}}
    \caption{Examples of two datasets. Top: Matterport3D; Bottom: Structured3D.}
    \label{fig:supp_dataset}
\end{figure*}
\begin{table*}[!t]
    \centering
    \caption{\textbf{Quantitative comparison with ERP-based baselines on Matterport3D test dataset}. For most conditions, we set the channel number $C$ to 64. Especially, we set $C$ to 256 for ResNet-50$^*$ and to 128 for Swin-B$^*$. \textbf{Bold} indicates performance improvement. \colorbox{red!10}{\textcolor{red}{\textbf{Red}}} indicates performance decline.}
    \label{tab:supp_comparison_erp}
    \resizebox{1\textwidth}{!}{ 
    \begin{tabular}{c|c|c|c|c|c|c|c|c|c|c|c|c}
    \toprule[1pt]
    \multirow{2}*{Backbone}& \multirow{2}*{Method} &\multirow{2}*{$\#$Params (M)}&\multirow{2}*{$\#$FLOPs (G)}& \multicolumn{4}{c|}{Depth} & \multicolumn{3}{c|}{Normal} & \multicolumn{2}{c}{Segmentation} \\
    \cmidrule{5-13}
    & & & & Abs Rel $\downarrow$ & RMSE $\downarrow$ &$\delta_1 (\%)$ $\uparrow$ &$\delta_2 (\%)$ $\uparrow$& Mean $\downarrow$& $\alpha_{11.25^\circ} (\%) \uparrow$& $\alpha_{22.5^\circ} (\%) \uparrow$&PixAcc \% $\uparrow$&mIoU \%  $\uparrow$\\
    \midrule[1pt]
    \multirow{3}*{Res-50$^*$}&Equi-Base& 34.70& 143.32& 0.1316& 0.4775& 87.50& 95.38& 5.8910& 87.88& 91.98& 94.05&74.96\\
    &Ours&38.25& 167.23& 0.1234 & 0.4279 & 88.56& 95.85 & 5.6516 & 88.23 &92.35 &93.91 & 74.24\\
    &\cellcolor{gray!40}$\varDelta$ &\cellcolor{gray!40}\textbf{+3.55}& \cellcolor{gray!40}\textbf{+23.91}& \cellcolor{gray!40}\textbf{-6.23$\%$}&\cellcolor{gray!40}\textbf{-10.39$\%$}& \cellcolor{gray!40}\textbf{+1.06$\%$}& \cellcolor{gray!40}\textbf{-0.47$\%$}& \cellcolor{gray!40}\textbf{-4.06$\%$}& \cellcolor{gray!40}\textbf{+0.35$\%$}& \cellcolor{gray!40}\textbf{+0.37$\%$}&
    \cellcolor{gray!40}-0.14$\%$& \cellcolor{gray!40}-0.72$\%$\\
    \midrule
    \multirow{3}*{Eff-B5}&Equi-Base& 37.22& 106.92& 0.1228 & 0.4136&88.55&95.74& 4.3532&90.04 & 94.08 &94.93&77.24\\
    &Ours&38.24 & 112.98& 0.1170 & 0.4044& 89.76& 95.89&4.3686 & 90.12& 93.96 & 94.56 &  75.86 \\
    &\cellcolor{gray!40}$\varDelta$ &\cellcolor{gray!40}\textbf{+1.02}& \cellcolor{gray!40}\textbf{+6.06}& \cellcolor{gray!40}\textbf{-4.72$\%$}&\cellcolor{gray!40}\textbf{-2.22$\%$}& \cellcolor{gray!40}\textbf{+1.21$\%$} &\cellcolor{gray!40}\textbf{+0.15$\%$}& \cellcolor{gray!40}+0.35$\%$& \cellcolor{gray!40}\textbf{+0.08$\%$}& \cellcolor{gray!40}-0.12$\%$& 
    \cellcolor{gray!40}-0.37$\%$& \cellcolor{gray!40}-1.38$\%$\\
    \midrule
    \multirow{3}*{Swin-B$^*$}&Equi-Base &95.24&244.75 & 0.1436 &0.5094&86.36&94.21& 5.6467& 87.96& 92.07 & 90.88&67.60\\
    &Ours&98.79 & 268.66&0.1483 &0.5178&85.58& 94.55&5.7334& 88.04 &92.07 & 91.16&67.98\\
    &\cellcolor{gray!40}$\varDelta$ &\cellcolor{gray!40}\textbf{+3.55}& \cellcolor{gray!40}\textbf{+23.91}& \cellcolor{gray!40}+3.27$\%$&\cellcolor{gray!40}+1.65$\%$& \cellcolor{gray!40}-0.78$\%$& \cellcolor{gray!40}\textbf{+0.34$\%$}& \cellcolor{gray!40}+1.53$\%$& \cellcolor{gray!40}\textbf{+0.08$\%$}& \cellcolor{gray!40}+0.00$\%$& \cellcolor{gray!40}\textbf{+0.28$\%$}& \cellcolor{gray!40}\textbf{+0.38$\%$} \\
    \bottomrule[1pt]
    \end{tabular}}
\end{table*}
\section{Datasets and Evaluation Metrics}
\label{sec:supp_dataset}
\noindent \textbf{Datasets.} Compared to planar images, there are fewer multi-task datasets based on 360$^\circ$ images. We have selected two large-scale datasets: Matterport3D~\cite{Chang2017Matterport3DLF} and Structured3D~\cite{Zheng2019Structured3DAL}. Matterport3D is a real-world datasets, while Structured3D is a synthetic dataset. The statics of two datasets are listed in Tab.~\ref{tab:sup_dataset}. Matterport3D totally contains 10,800 ERP images, including 7,829 samples for training, 957 samples for validation, and 2014 samples for testing. For Structured3D, a recently proposed large-scale synthetic dataset, it includes 18343 training samples, 1776 validation samplings, and 1697 testing samples. During training and testing, we set the resolutions of all samples as $512 \times 1024$. The examples of two datasets are illustrated in Fig.~\ref{fig:supp_dataset}.

\noindent \textbf{Evaluation metrics.} For the depth evaluation metrics, we adopted widely used metrics from 360 depth estimation methods, such as BiFuse~\cite{Wang2020BiFuseM3}, UniFuse~\cite{Jiang2021UniFuseUF}, and HRDFuse~\cite{Ai2023HRDFuseM3}. These metrics include three error metrics: absolute relative error (Abs Rel), squared relative error (Sq Rel), root mean squared error (RMSE), and three threshold percentage $\delta<\alpha^{t}$ $(\alpha=1.25, t=1,2,3)$, denoted as $\delta_{t}$.  For the surface normal evaluation metrics, we follow the planar surface normal estimation methods~\cite{invpt2022,Bae2021EstimatingAE} to employ the common metrics, \ie, the mean angular error (Mean) and the percentage of pixel vectors within a preset angle $\alpha_a$, where $a \in \{ 11.25^\circ,22.5^\circ,30^\circ\}$. The semantic segmentation task is evaluated by the two metrics, mean Intersection over Union (mIoU) and mean pixel accuracy (PixAcc). Particularly noteworthy is that, for all metric computations, we employed the common practice of considering only valid pixels~\cite{Jiang2021UniFuseUF,Ai2023HRDFuseM3}\footnote{\url{https://github.com/alibaba/UniFuse-Unidirectional-Fusion/}}.
\begin{figure*}[ht]
    \centering
    \resizebox{0.95\linewidth}{!}{ \includegraphics{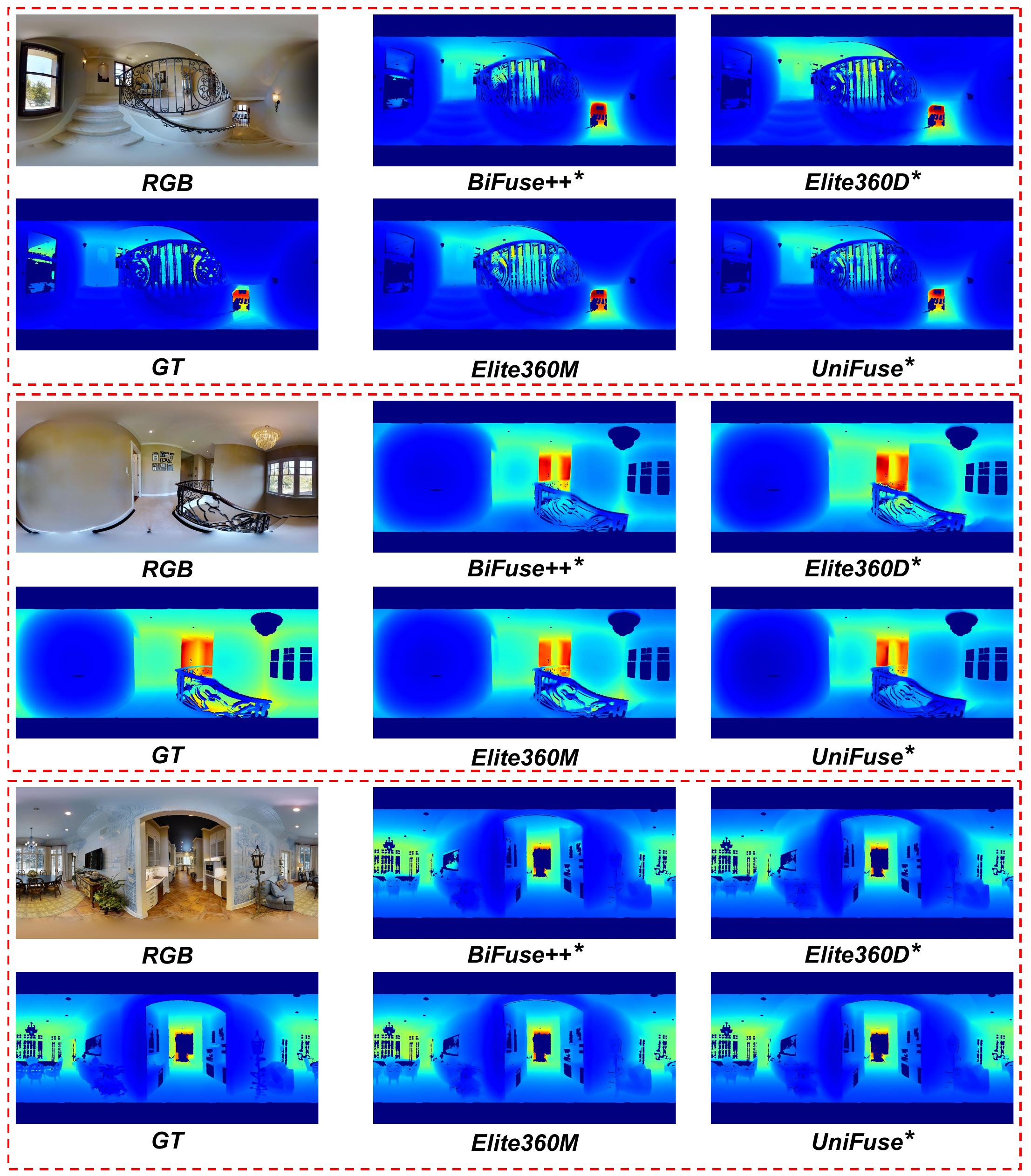}}
    \caption{Visual comparisons of depth estimation with single-task learning methods on Matterport3D dataset (All methods are with ResNet-34 as the backbone).}
    \label{fig:supp_single_depth}
\end{figure*}
\begin{figure*}[ht]
    \centering
    \resizebox{0.95\linewidth}{!}{ \includegraphics{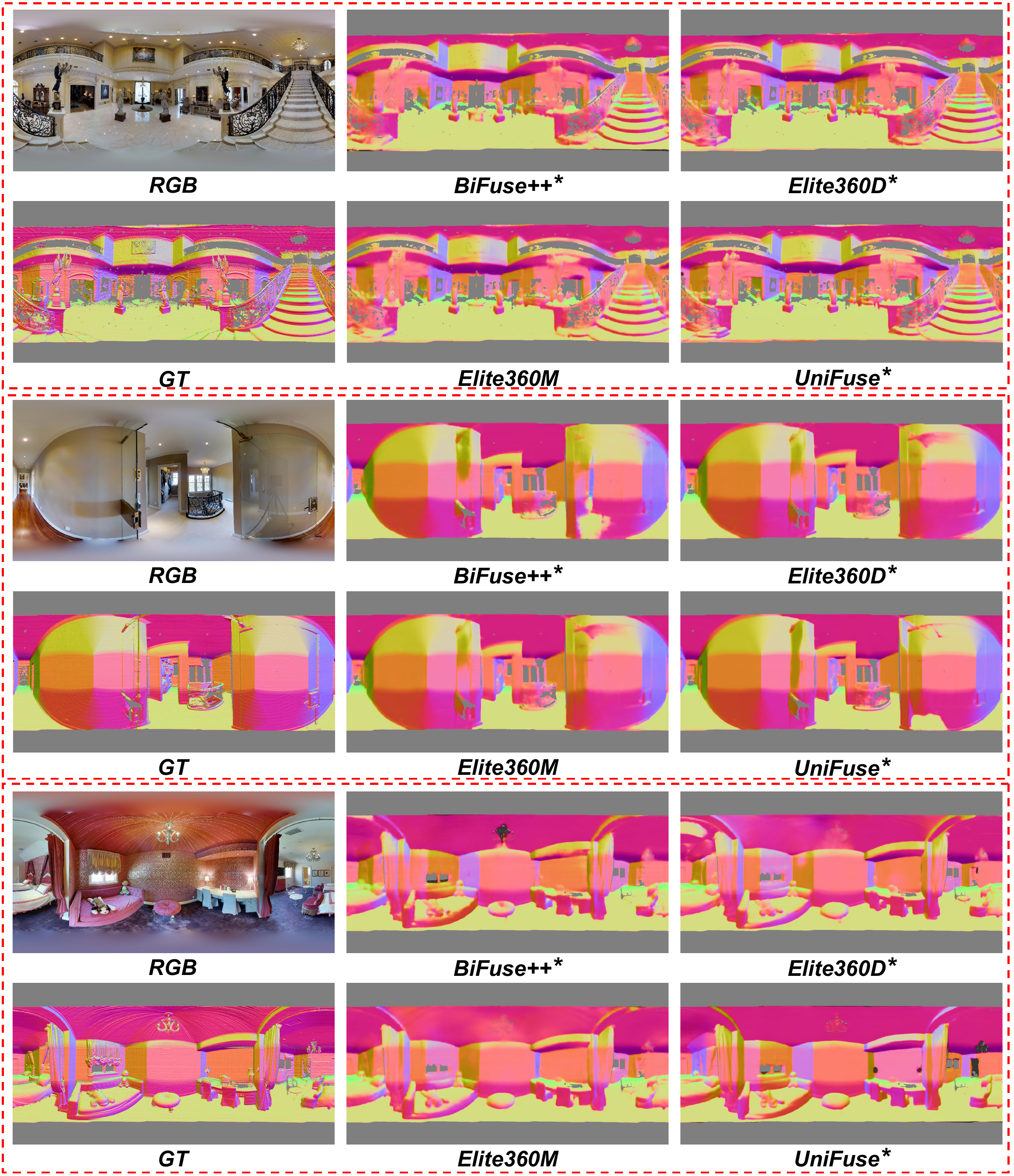}}
    \caption{Visual comparisons of surface normal estimation with single-task learning methods on Matterport3D dataset (All methods are with ResNet-34 as the backbone).}
    \label{fig:supp_single_normal}
\end{figure*}
\begin{figure*}[ht]
    \centering
    \resizebox{0.95\linewidth}{!}{ \includegraphics{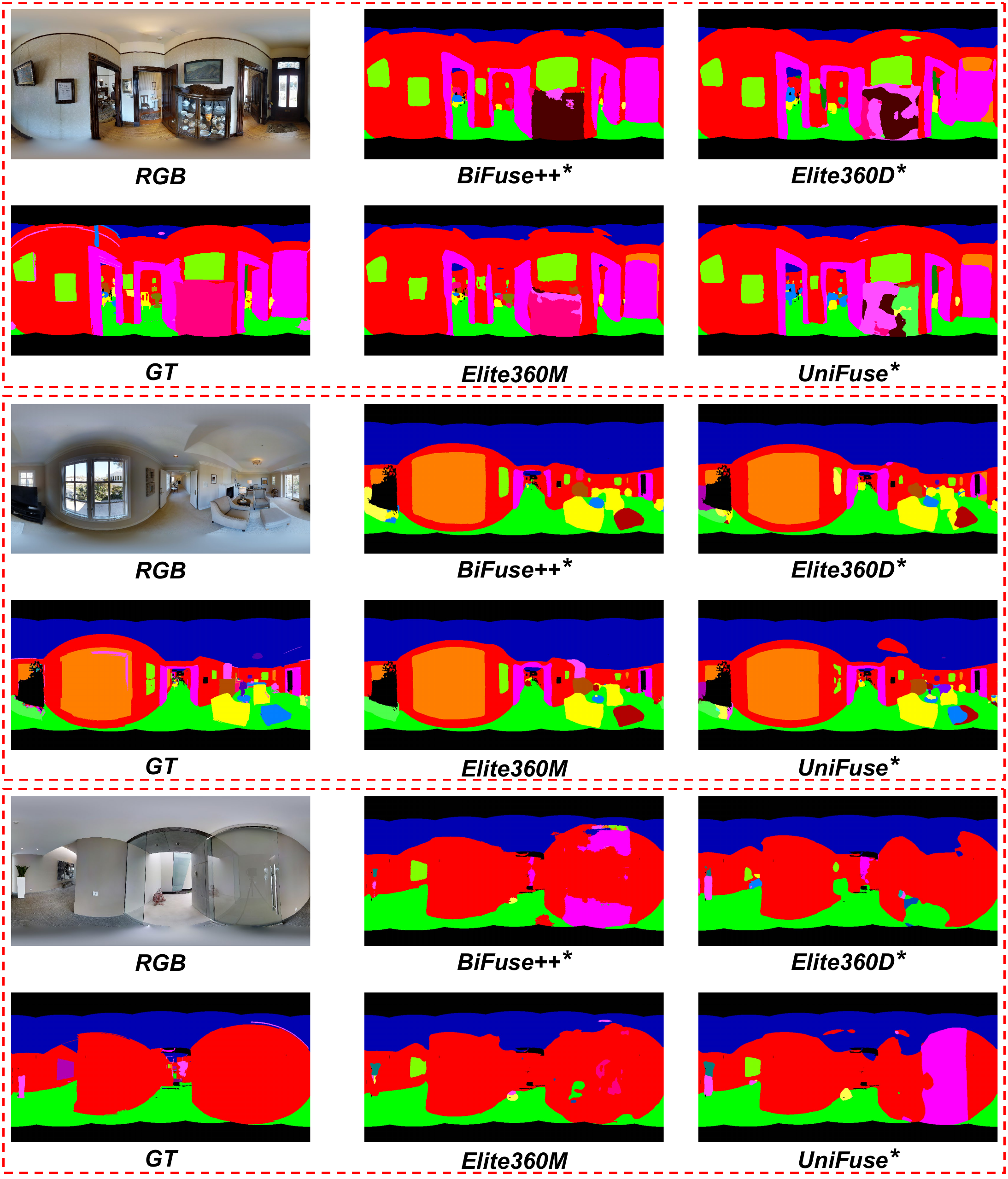}}
    \caption{Visual comparisons of semantic segmentation with single-task learning methods on Matterport3D dataset (All methods are with ResNet-34 as the backbone).}
    \label{fig:supp_single_seg}
\end{figure*}

\section{Training Details}
\label{sec:supp_training_detail}
Considering that existing methods for different tasks have varied training settings, \eg, optimizer, learning rate scheduler, final depth prediction, loss function and the number of training epochs. Furthermore, there is the blank in end-to-end 360$^\circ$ multi-task learning. Therefore, in this study, for a fair comparison and to better assess the effectiveness of different methods, we re-trained all methods using their official codes and standardized training settings. Specifically, for the optimizer, we used the Adam optimizer~\cite{Kingma2014AdamAM} with default settings. Meanwhile, we also avoid any learning rate scheduler. For the final depth prediction, We first utilized the sigmoid function to output depth probabilities, which are then multiplied by the maximum depth value to obtain the final depth values. For the surface normal prediction, we employ a normalization strategy~\cite{Yin2023Metric3DTZ} to normalize the final output. To ensure convergence for each method, we trained all methods for 200 epochs on Matterport3D dataset and 60 epochs on Structured3D dataset. For the ERP-based baseline, we directly removed the ICOSAP feature extraction and B2F module in Fig.~\ref{fig:supp_nework_achitecture}. We employed the encoder-decoder network with the CoCo module to directly accomplish three tasks based on the ERP input.

\section{Additional Quantitative and Qualitative Results}
\label{sec:supp_comparison_results}
\noindent \textbf{More quantitative results.} In Table.~\ref{tab:supp_comparison_erp}, we present more comparison results with the ERP baseline on Structured3D dataset, while employing the various encoder backbones. We observe that with a CNN-based encoder backbone (ResNet-50 and EfficientNet-B5), our Elite360M significantly outperforms ERP baselines in the depth estimation task and exceeds them in surface normal estimation, while incurring only minimal computational costs. However, for the semantic segmentation task, Elite360M slightly underperforms compared to ERP baselines. This may be attributed to the structural details of the ICOSAP point feature set, which introduce overly detailed semantic information, resulting in unevenness in some regions. Besides, while taking Swin-B as the encoder backbone, it can be observed that the enhancement from bi-projection fusion is slight and even results in decreased performance in depth estimation and surface normalization estimation. Considering the performance of bi-projection fusion in Elite360D~\cite{Ai2024Elite360DTE}, we believe this decline in accuracy is due to the imbalance among different tasks during multi-task learning, particularly impacting the transformer architecture.
\begin{figure*}[ht]
    \centering
    \resizebox{1\linewidth}{!}{ \includegraphics{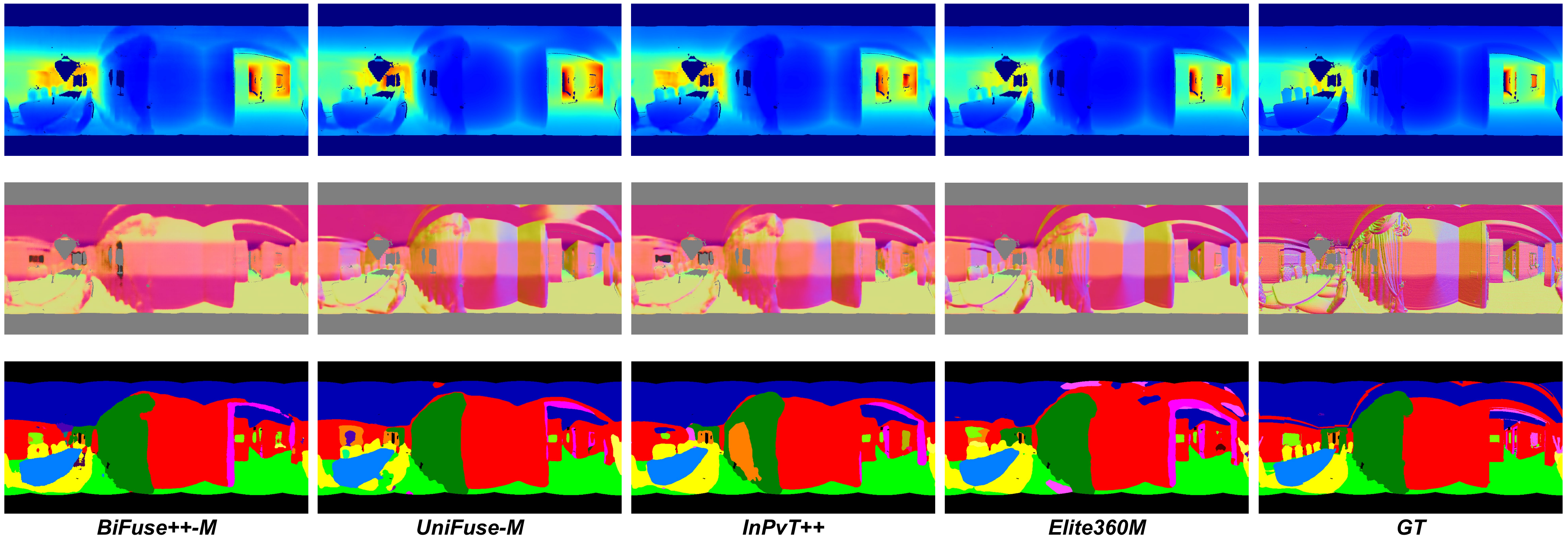}}
    \caption{Visual comparisons with multi-task learning baselines on Matterport3D dataset (All bi-projection based methods are with ResNet-34 as the backbone).}
    \label{fig:supp_multi1}
\end{figure*}
\begin{figure*}[ht]
    \centering
    \resizebox{1\linewidth}{!}{ \includegraphics{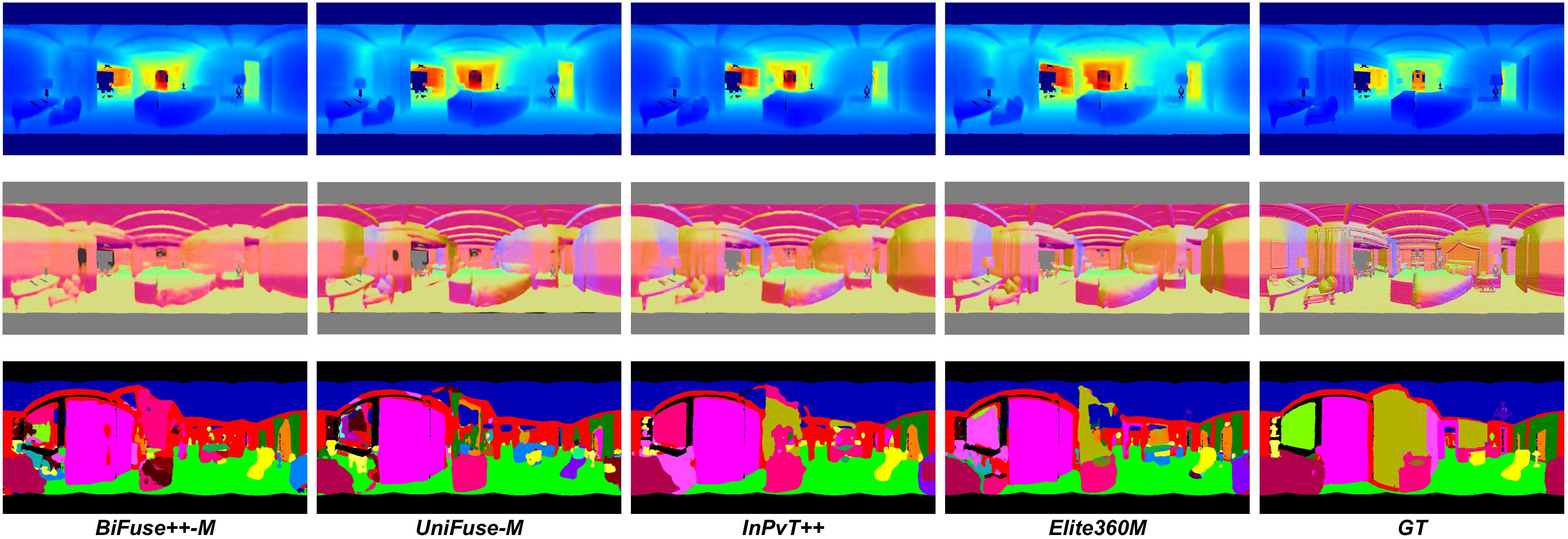}}
    \caption{Visual comparisons with multi-task learning baselines on Matterport3D dataset (All bi-projection based methods are with ResNet-34 as the backbone).}
    \label{fig:supp_multi2}
\end{figure*}

\noindent \textbf{More qualitative results.}
In Fig.~\ref{fig:supp_single_depth}, Fig.~\ref{fig:supp_single_seg} and Fig.~\ref{fig:supp_single_normal}, we present visual comparisons for depth estimation, surface normal estimation, and semantic segmentation, respectively, between our Elite360M and various bi-projection-based single-task learning methods on the Matterport3D dataset. Notably, even when benchmarked against supervised single-task methods, Elite360M consistently delivers comparable or superior results.

Furthermore, in Fig.~\ref{fig:supp_multi1} and Fig.~\ref{fig:supp_multi2}, we showcase the visual results comparing our Elite360M with multi-task learning baselines. It is evident that Elite360M predicts more accurate outcomes at a significantly low computational cost. Meanwhile, 

\clearpage
{
\bibliographystyle{IEEEtran}
 \bibliography{egbib}
}